\pdfoutput=1

\documentclass[11pt]{article}

\usepackage{acl}
\usepackage{times}
\usepackage{latexsym}
\usepackage{amsmath,amssymb}
\usepackage{mathabx}

\usepackage{comment}
\usepackage{varwidth}
\usepackage{bbm}
\usepackage{pifont}
\usepackage{mathbbol}
\usepackage[edges]{forest}
\usepackage{rotating}
\usepackage[utf8]{inputenc}
\usepackage{times}
\usepackage{pgfplots}
\usepackage{tikz}
\pgfplotsset{compat=newest}
\usepackage{latexsym}
\usepackage{rotating,tabularx,siunitx}
\definecolor{cYellow}{RGB}{255,255,3}
\definecolor{cBlue}{RGB}{69,123,157}
\definecolor{cRed}{RGB}{231,56,71}
\definecolor{cRed_1}{RGB}{191,30,46}
\definecolor{cGray}{RGB}{168,218,219}
\definecolor{cBlue_2}{RGB}{5,48,97}
\definecolor{cBlue_1}{RGB}{115,186,214}
\definecolor{cBlue_3}{RGB}{13,76,109}
\definecolor{cBlue_4}{RGB}{64,121,160}
\definecolor{cBlue_5}{RGB}{64,121,160}
\definecolor{cOrange}{RGB}{250,134,0}
\definecolor{cBlue_6}{RGB}{13,76,109}
\definecolor{cBlue_7}{RGB}{16,106,130}
\definecolor{cBlue_8}{RGB}{19,136,160}
\definecolor{cBlue_9}{RGB}{115,184,214}

\definecolor{above90}{RGB}{67,160,71} 
\definecolor{above80}{RGB}{13,76,109}
\definecolor{above70}{RGB}{215,175,24}
\definecolor{above60}{RGB}{244,81,30}
\definecolor{below60}{RGB}{90,90,90}

\definecolor{hidden-red}{RGB}{205, 44, 36}
\definecolor{hidden-blue}{RGB}{194,232,247}
\definecolor{hidden-orange}{RGB}{243,202,120}
\definecolor{hidden-green}{RGB}{34,139,34}
\definecolor{hidden-pink}{RGB}{245,235,237}
\definecolor{hidden-black}{RGB}{20,68,106}

\usepackage{float}
\usetikzlibrary{plotmarks}
\usepackage{graphicx}
\usepackage{subfigure}
\usepackage{caption}
\usepackage{mathtools}
\usepackage{booktabs}
\usepackage{multirow}
\usepackage{ragged2e}
\usepackage{hyperref}
\usepackage{multicol}
\usepackage{siunitx}
\usepackage{adjustbox}
\usepackage{pifont}
\usepackage{tabularx}
\usepackage{enumitem}
\usepackage{pythonhighlight}
\usepackage{array, makecell, multirow}
\newcolumntype{P}[1]{>{\centering\arraybackslash}p{#1}}
\usepackage{siunitx}
\usepackage{soul}
\usepackage{xcolor}
\usepackage{colortbl}
\sethlcolor{cYellow}
\usepackage{tcolorbox}
\usepackage{amsmath}
\usepackage{amssymb}
\usepackage{pifont}
\definecolor{coffee1}{RGB}{111,78,55}    
\definecolor{coffee2}{RGB}{212,163,115}  
\definecolor{coffee3}{RGB}{139,90,43}    
\definecolor{coffee4}{RGB}{222,184,135}  
\newcolumntype{R}[1]{>{\RaggedLeft\arraybackslash}p{#1}}
\newcolumntype{L}[1]{>{\RaggedRight\arraybackslash}p{#1}}
\tcbuselibrary{skins, breakable, theorems}

\usepackage[T1]{fontenc}
\usepackage[utf8]{inputenc}

\usetikzlibrary{shapes.geometric}

\usepackage{microtype}
\usepackage{inconsolata}
\pgfplotsset{
/pgfplots/my legend/.style={
legend image code/.code={
    \node[star,star point ratio=2.25,rotate=36,minimum size=12pt,
          inner sep=0pt,draw=cBlue_4,solid,fill=cBlue_4,fill opacity=0.3] {};
   }
  }
}

\pgfplotsset{
/pgfplots/my medium/.style={
legend image code/.code={
    \node[fill opacity=0.7,text=cOrange] {\large $\odot$};
   }
  }
}
\pgfplotsset{
/pgfplots/my easy/.style={
legend image code/.code={
    \node[fill opacity=0.7,text=cBlue_3] {\large $\oplus$};
   }
  }
}

\pgfplotsset{
/pgfplots/my hard/.style={
legend image code/.code={
    \node[fill opacity=0.7,text=cRed_1] {\large $\otimes$};
   }
  }
}

\pgfdeclareplotmark{mystar}{
    \node[star,star point ratio=2.25,minimum size=12pt,rotate=36,
          inner sep=0pt,draw=cBlue_4,solid,fill=cBlue_4,fill opacity=0.3] {};
}
\pgfdeclareplotmark{my_easy}{
    \node[fill opacity=0.3,text=cBlue_3] {$\oplus$};
}
\pgfdeclareplotmark{my_medium}{
    \node[fill opacity=0.3,text=cOrange] {$\odot$};
}
\pgfdeclareplotmark{my_hard}{
    \node[fill opacity=0.3,text=cRed_1] {$\otimes$};
}
\usepackage{fdsymbol}

\title{Structuring the Unstructured: A Systematic Review of Text-to-Structure Generation for Agentic AI with a Universal Evaluation Framework}

\author{Zheye Deng, Chunkit Chan, Tianshi Zheng, Wei Fan, Weiqi Wang, Yangqiu Song\\
  Department of Computer Science and Engineering, HKUST, Hong Kong SAR, China\\
  \texttt{zdengah@cse.ust.hk}\\
}

\begin{document}

\maketitle
\begin{abstract}

The evolution of AI systems toward agentic operation and context-aware retrieval necessitates transforming unstructured text into structured formats like tables, knowledge graphs, and charts. While such conversions enable critical applications from summarization to data mining, current research lacks a comprehensive synthesis of methodologies, datasets, and metrics. 
This systematic review examines text-to-structure techniques and the encountered challenges, evaluates current datasets and assessment criteria, and outlines potential directions for future research. We also introduce a universal evaluation framework for structured outputs, establishing text-to-structure as foundational infrastructure for next-generation AI systems.

\end{abstract}

\section{Introduction}

The rapid growth of the agentic AI systems is redefining the paradigm of information processing, where autonomous agents must dynamically acquire, synthesize, and act upon structured knowledge extracted from textual sources~\cite{DBLP:journals/corr/abs-2501-09136,DBLP:journals/corr/abs-2502-13025}. This agentic revolution creates dual demands for structured knowledge representation: (1) enabling dynamic knowledge grounding during multi-step agentic reasoning, and (2) serving as curated retrieval sources for Retrieval-Augmented Generation (RAG) pipelines~\cite{DBLP:journals/corr/abs-2410-09713,DBLP:journals/corr/abs-2502-16866,DBLP:journals/corr/abs-2501-09136, li2025agenticragdeepreasoning, DBLP:journals/corr/abs-2506-18959}.

Complex structures, such as tables and graphs, play a crucial role in conveying information, as they can intuitively display data relationships and temporal trends~\cite{DBLP:journals/corr/abs-2404-07738,DBLP:conf/icml/HuangVLL24,DBLP:journals/corr/abs-2409-05556,DBLP:journals/corr/abs-2410-13185}, while preserving hierarchical dependencies, which are vital for enhancing RAG reliability~\cite{DBLP:journals/corr/abs-2404-16130,DBLP:conf/emnlp/ZhuangZCYLHLR0Z24,DBLP:journals/corr/abs-2410-08815,DBLP:conf/emnlp/WangRLZLW24}. These structures enable both comprehension of complex information and machine-friendly representations for downstream processing~\cite{DBLP:conf/acl/JainMP24,DBLP:conf/chi/ReifQWK24}.

\begin{figure}[t]
    \centering
    \includegraphics[width=\linewidth]{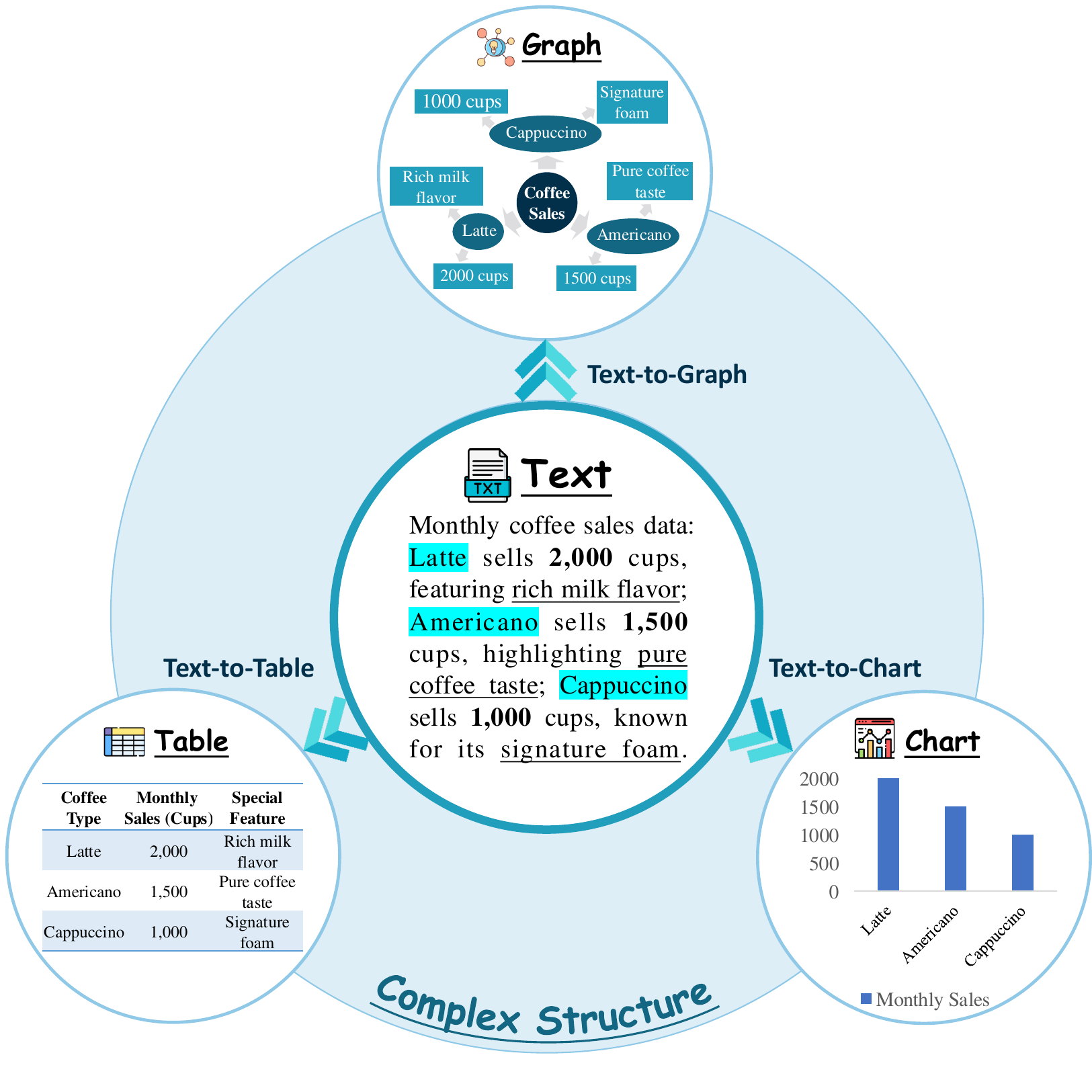}
    \vspace{-20pt}
    \caption{An example of text-to-complex structure conversion. The center represents the text, while the surrounding elements constitute the complex structures. Based on this text, three types of complex structures -- table, graph, and chart -- can be generated respectively.}
    \label{fig:intro}
    \vspace{-20pt}
\end{figure}

Traditional NLP methods have demonstrated proficient performance in extracting simple entities and relationships~\cite{DBLP:conf/conll/SangM03,DBLP:conf/acl/YaoYLHLLLHZS19}, but their efficacy is limited when confronted with complex structured knowledge implicit in the text. The emergence of Large Language Models (LLMs) enhances the capabilities of language models in capturing complex semantics and long-range dependencies, crucial for supporting agentic systems' dynamic knowledge requirements and RAG's structured retrieval needs. This advancement drives the growth of research in complex structured information extraction. Currently, the most common forms of structured output include tables, knowledge graphs, mind maps, and charts, which are crucial in downstream tasks such as RAG-enhanced question-answering and agentic data analysis~\cite{DBLP:journals/corr/abs-2406-03618,DBLP:journals/corr/abs-2503-13269}. However, this task also poses considerable challenges: complex structured information is represented in diverse forms in the text, and accurately extracting this information requires a deep semantic understanding of the text, including identifying implicit relationships and reasoning. Moreover, there is a lack of task-specific data tailored for training and evaluation. To address these issues, researchers have explored various methods to enhance model performance and constructed several high-quality datasets for benchmarking.

\begin{figure}[!t]
    \centering
    \includegraphics[width=\linewidth]{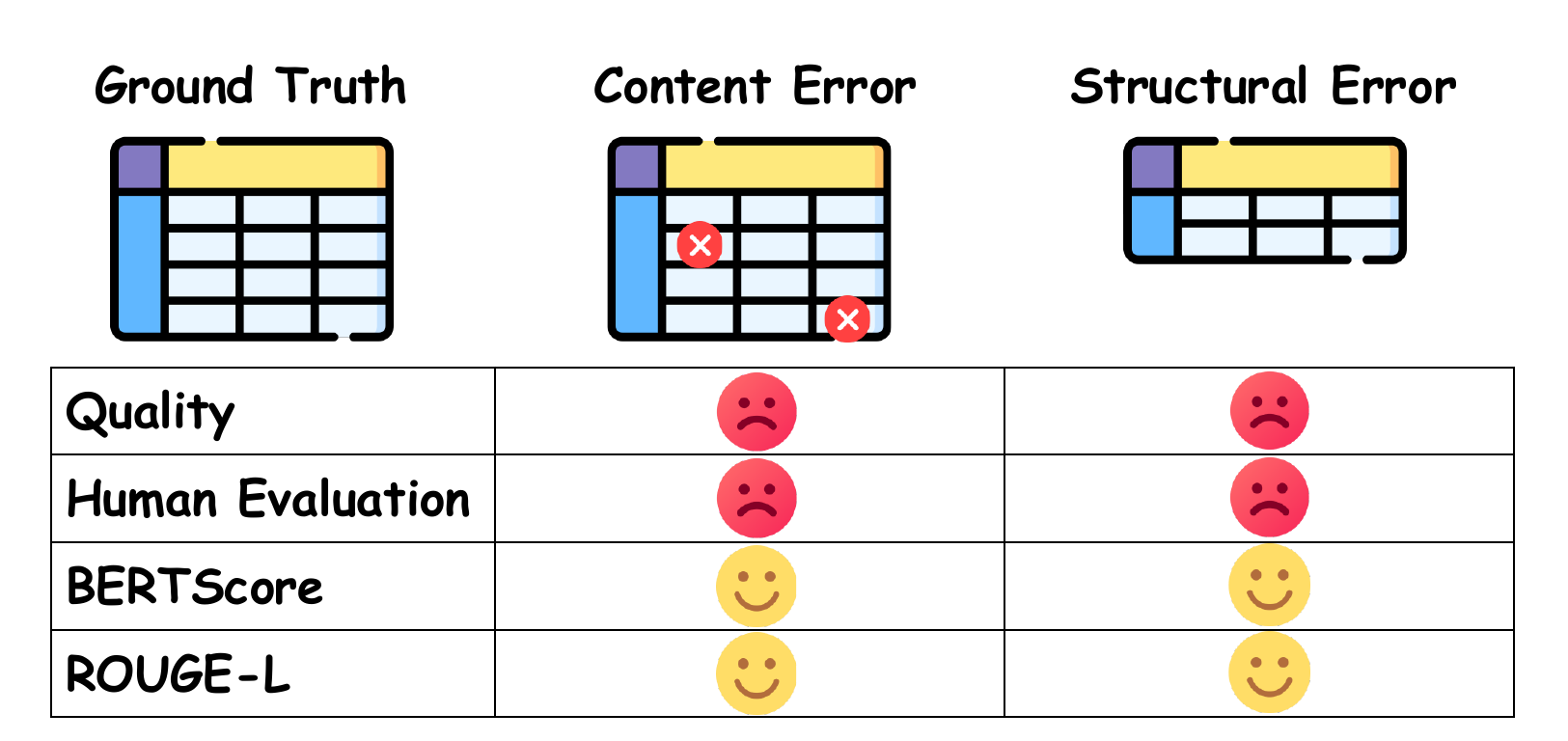}
    \caption{Misalignment between traditional metrics and true quality in structured output evaluation.}
    \label{fig:misalign}
    \vspace{-10pt}
\end{figure}

Notwithstanding the significance of this task, the field currently lacks a comprehensive survey to summarize the existing research progress. This gap becomes more critical with the rise of agentic systems that demand structured knowledge representations~\cite{DBLP:journals/corr/abs-2410-09713}. Specifically, two key challenges persist: (1) the baseline models and benchmark datasets used for comparison vary significantly across studies; (2) current evaluation systems remain constrained by traditional metrics that systematically misalign with human judgment when assessing structured outputs (as demonstrated in Figure~\ref{fig:misalign}).
To bridge these gaps, our contributions are threefold:
\begin{itemize}[itemsep=1pt, parsep=0pt, topsep=1pt, partopsep=0pt,leftmargin=20pt]
    \item We conduct a systematic synthesis and critical analysis of current methods (\S\ref{sec:methods}), datasets (\S\ref{sec:datasets}), and evaluation metrics (\S\ref{sec:metrics}).
    \item We establish a novel Text-to-Structure (T2S) benchmark with a universal framework for structured output evaluation (\S\ref{sec:advie}).
    \item Extensive experimental results show that our framework significantly outperforms traditional metrics, providing actionable resources.
\end{itemize}

\section{Task Formulation} \label{sec:tasks}

As shown in Figure~\ref{fig:intro}, we categorize Text-to-Structure generation into three common types: tables, graphs, and charts. For any task, the input should consist of a textual passage with $N$ tokens, denoted as {\small $\mathcal{T}=[t_1,t_2,\dots,t_N]$}, and an instruction text with $M$ tokens, denoted as {\small $\mathcal{I}=[i_1,i_2,\dots,i_M]$} as constituent elements. We then provide formal definitions and detailed explanations for each category to demonstrate their distinct characteristics. 

\subsection{Table Generation} \label{sec:table_gen}
We define the table $\boldsymbol{T}$ as: $\boldsymbol{T}=(\mathcal{R,C,E},Capt)$, where $\mathcal{R}=\{r_1,r_2,\dots,r_m\}$ is an ordered set of $m$ rows, $\mathcal{C}=\{c_1,c_2,\dots,c_n\}$ is an ordered set of $n$ columns, $\mathcal{E}: \mathcal{R}\times \mathcal{C}\rightarrow \mathcal{D}$ is a function that assigns a data entry to each cell in the table, and $Capt$ denotes the caption of the table. It is allowed to generate multiple tables at the same time~\cite{DBLP:conf/acl/0001ZL22,DBLP:conf/naacl/TangZPZZCG24,DBLP:conf/acl/JainMP24}.

\tikzstyle{my-box}=[
    rectangle,
    draw=hidden-black,
    rounded corners,
    text opacity=1,
    minimum height=1.5em,
    minimum width=5em,
    inner sep=2pt,
    align=center,
    fill opacity=.5,
]
\tikzstyle{leaf}=[
    my-box, 
    minimum height=1.5em,
    fill=hidden-blue!90, 
    text=black,
    align=left,
    font=\normalsize,
    inner xsep=2pt,
    inner ysep=4pt,
]
\begin{figure*}[t]
    \centering
    \resizebox{\textwidth}{!}{
        \begin{forest}
            forked edges,
            for tree={
                grow=east,
                reversed=true,
                anchor=base west,
                parent anchor=east,
                child anchor=west,
                base=left,
                font=\large,
                rectangle,
                draw=hidden-black,
                rounded corners,
                align=left,
                minimum width=4em,
                edge+={darkgray, line width=1pt},
                s sep=3pt,
                inner xsep=2pt,
                inner ysep=3pt,
                line width=0.8pt,
                ver/.style={rotate=90, child anchor=north, parent anchor=south, anchor=center},
            },
            where level=1{text width=5.2em,font=\normalsize,}{},
            where level=2{text width=7.4em,font=\normalsize,}{},
            where level=3{text width=4.85em,font=\normalsize,}{},
            where level=4{text width=12em,font=\normalsize,}{},
            [
                \ Complex Structured Output Generation \ , ver
                [
                    \parbox{5.2em}{Tasks \hfill (\S \ref{sec:tasks})}
                    [
                        Table\\Generation (\S \ref{sec:table_gen})
                        [
                        \parbox{35em}{\citet{DBLP:conf/acl/0001ZL22,DBLP:conf/acl/LiWSZWS23,DBLP:conf/naacl/TangZPZZCG24,DBLP:conf/emnlp/DengC00FZYS24}; \citet{DBLP:conf/acl/JainMP24,DBLP:conf/emnlp/JiangLZMCC24,DBLP:conf/emnlp/RamuGB24,DBLP:conf/emnlp/NewmanLNSFKWCL24}}
                            , leaf, text width=35em, align=left
                        ]
                    ]
                    [
                        Graph\\Generation (\S \ref{sec:graph_gen})
                        [
                            Knowledge\\Graph
                            [
                            \parbox{28.5em}{\citet{DBLP:conf/emnlp/MelnykDD22}; \citet{DBLP:conf/semweb/MihindukulasooriyaTEL23}; \citet{DBLP:journals/corr/abs-2307-01128}; \citet{DBLP:journals/information/GopalakrishnanC23}; \citet{DBLP:journals/corr/abs-2305-11527}; \citet{DBLP:journals/biodb/MirandaEscaladaMLEGPVK23}; \citet{DBLP:conf/bigdataconf/ChenB23}; \citet{DBLP:journals/corr/abs-2401-04507}; \citet{DBLP:journals/corr/abs-2410-17600};
                                \citet{huang2024llmsgoodgraphjudger}}
                                , leaf, text width=28.5em
                            ]
                        ]
                        [
                            Mind Map
                            [
                                \parbox{28.5em}{
                                \citet{DBLP:journals/mta/ElhoseinyE16}; \citet{DBLP:conf/ijcai/LiuCCWQH19}; \citet{DBLP:conf/ijcai/WeiGWS19}; \citet{DBLP:conf/emnlp/HuGZGS21}; \citet{DBLP:journals/nca/MohammedF23}; \citet{10420315}; \citet{DBLP:conf/acl/JainMP24}; \citet{DBLP:conf/aaai/ZhangHBZ24}}
                                , leaf, text width=28.5em
                            ]
                        ]
                    ]
                    [
                        Chart\\Generation (\S \ref{sec:chart_gen})
                        [
                        \parbox{35em}{\citet{DBLP:conf/pakdd/RashidJHRZMMS22}; \citet{DBLP:journals/corr/abs-2311-01920}; \citet{DBLP:journals/corr/abs-2311-16483}; \citet{DBLP:journals/corr/abs-2410-14331}; \citet{DBLP:journals/corr/abs-2404-04854}; \citet{DBLP:conf/emnlp/ZadehKKK24}; \citet{DBLP:journals/tvcg/XiaoHLYZ24}; \citet{Yildirim2024NLPIF}}
                            , leaf, text width=35em
                        ]
                    ]
                ]
                [
                    \parbox{5.2em}{Methods\hfill(\S \ref{sec:methods})}
                    [
                        \parbox{7.4em}{Fine Tuning\hfill (\S \ref{sec:methods:fine-tune})}
                        [
                            \parbox{35em}{
                            \citet{DBLP:conf/acl/0001ZL22}; \citet{DBLP:conf/acl/LiWSZWS23};  {Odie}~\cite{DBLP:conf/emnlp/Jiao0LZOJ023};
                            gTBLS~\cite{DBLP:journals/corr/abs-2403-14457}; STable~\cite{DBLP:conf/eacl/PietruszkaTBDNSJG24}; ~\citet{DBLP:conf/naacl/TangZPZZCG24};~\citet{DBLP:conf/emnlp/ZadehKKK24}
                            } 
                            , leaf, text width=35em
                        ]
                    ]
                    [
                        \parbox{7.4em}{Prompting\hfill (\S \ref{sec:methods:prompting})}
                        [
                            \parbox{35em}{
                            {StructSum}~\cite{DBLP:conf/acl/JainMP24};~\citet{DBLP:conf/naacl/TangZPZZCG24};~\citet{huang2024llmsgoodgraphjudger}}
                            , leaf, text width=35em
                        ]
                    ]
                    [
                        \parbox{7.4em}{Hybrid\\Pipelines\hfill (\S \ref{sec:methods:hybrid})}
                        [
                        \parbox{35em}{
                            \citet{DBLP:conf/pakdd/RashidJHRZMMS22}; KG2Instruction~\cite{DBLP:journals/corr/abs-2305-11527}; T\textsuperscript{3}~\cite{DBLP:conf/emnlp/DengC00FZYS24}; TKGT~\cite{DBLP:conf/emnlp/JiangLZMCC24}; 
                            GraphJudger~\cite{huang2024llmsgoodgraphjudger}; Graphusion~\cite{DBLP:journals/corr/abs-2410-17600};~\citet{DBLP:journals/corr/abs-2410-14331}}
                            , leaf, text width=35em
                        ]
                    ]
                ]
            ]
        \end{forest}
    }
    \vspace{-15pt}
    \caption{Taxonomy of representative works in text-to-structured generation categorized by tasks and methods.}
    \label{fig:taxonomy}
    \vspace{-15pt}
\end{figure*}
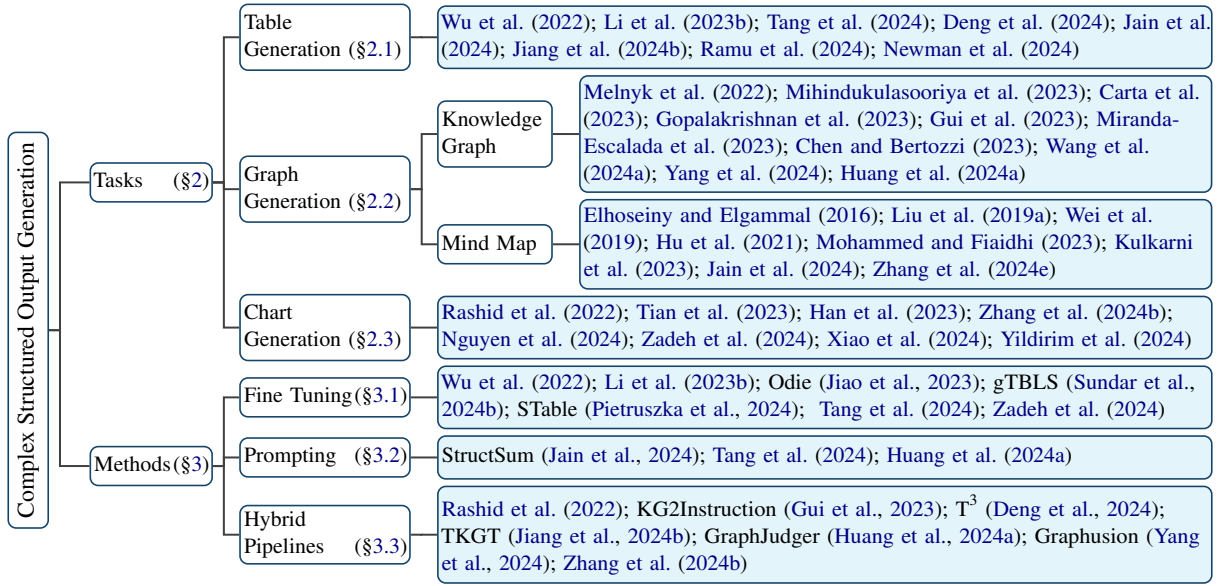

\subsection{Graph Generation} \label{sec:graph_gen}
A graph can be considered an extension of a table due to its greater flexibility and schema-less nature. To model more complex structures, we extend the definition of graphs to include attributes and semantics: $\boldsymbol{G}=(\mathcal{V},\mathcal{E},\mathcal{A},\mathcal{L})$, where $\mathcal{V}$ is a set of nodes representing entities or concepts, $\mathcal{E}\in \mathcal{V}\times \mathcal{V}$ is a set of edges representing relations between nodes, $\mathcal{A}: \mathcal{V} \cup\mathcal{E} \rightarrow \mathcal{P}$ is an attribute function assigning properties to nodes and edges, $\mathcal{L}: \mathcal{V} \cup \mathcal{E} \rightarrow \Sigma^*$ is a semantic function mapping nodes and edges to contextual meanings. The framework unifies graph-based structures, such as knowledge graphs and mind maps~\cite{DBLP:conf/ijcai/WeiGWS19,DBLP:conf/emnlp/HuGZGS21}. See Appendix~\ref{app:xxxx} for more details.

\subsection{Chart Generation} \label{sec:chart_gen}
A chart can be considered a visualization that goes beyond the structure of a graph. The input may include data-specific contexts, such as tables or data attributes (e.g., data types, ranges, etc.)~\cite{DBLP:journals/corr/abs-2311-01920}. The goal is to generate visualization charts that feature the appropriate specifications, including the type of the chart, the visual encodings, and other relevant details~\cite{DBLP:journals/corr/abs-2410-14331}. 

\section{Methods} \label{sec:methods}

\subsection{Fine Tuning} \label{sec:methods:fine-tune}
Supervised Fine Tuning (SFT) is still the default approach to text-to-structure generation since it only necessitates paired \emph{text$\rightarrow$target} instances. Although intuitive, its performance differs across tasks because of variances in data needs, structural intricacy, and computational expenses. In text-to-table generation, sequence-to-sequence models with header-aware serialization~\cite{DBLP:conf/acl/0001ZL22,DBLP:conf/acl/LiWSZWS23} produce validated tables but need enormous annotated datasets, which restricts low-resource applicability. QA-style cell queries~\cite{DBLP:journals/corr/abs-2404-12580,DBLP:journals/corr/abs-2403-14457} decrease data requirements, while being challenged by intricate layouts. Instruction tuning with Chain-of-Thought prompting~\cite{DBLP:conf/nips/Wei0SBIXCLZ22} and LLM-synthesized pairs increases robustness but includes computational expense~\cite{DBLP:conf/emnlp/Jiao0LZOJ023,DBLP:conf/naacl/TangZPZZCG24}.
For text-to-KG generation, SFT tends to follow a two-step pipeline, entity recognition followed by relation extraction~\cite{DBLP:journals/corr/abs-2305-11527,DBLP:journals/corr/abs-2401-04507,DBLP:journals/corr/abs-2410-17600}, which enhances structure but has a risk of error propagation. In text-to-chart tasks, instruction tuning with RL-based feedback makes it more adaptable, although performance relies on the quality of the feedback, particularly for uncommon chart types~\cite{DBLP:journals/corr/abs-2402-12185,DBLP:conf/emnlp/ZadehKKK24}.

\subsection{Prompting} \label{sec:methods:prompting}
The prosperity of LLMs~\cite{DBLP:journals/corr/abs-2407-21783,openai2024gpt4,deepseekai2025deepseekr1incentivizingreasoningcapability} has triggered the evolution of diverse prompting methods for text-to-structure tasks~\cite{DBLP:conf/nips/Wei0SBIXCLZ22,DBLP:conf/emnlp/DuaG0G22, DBLP:conf/iclr/KhotTFF0CS23}, with each presenting trade-offs in accuracy, efficiency, and simplicity. Basic prompts such as ``{\small \texttt{generate tables/graphs/charts}}'' are quick but have poor results~\cite{DBLP:conf/acl/JainMP24,huang2024llmsgoodgraphjudger}. Incorporating schema constraints enhances structural fidelity at the expense of potentially reducing flexibility in tasks with implicit relations~\cite{DBLP:conf/naacl/TangZPZZCG24,DBLP:conf/emnlp/DengC00FZYS24,DBLP:conf/acl/JainMP24}. Chain-of-Thought prompts are also used to enhance reasoning and accuracy~\cite{DBLP:conf/acl/Wang0W23,DBLP:conf/naacl/TangZPZZCG24}. Decomposed prompting improves accuracy by breaking up input~\cite{DBLP:conf/iclr/KhotTFF0CS23,DBLP:conf/acl/JainMP24}, but can lose global context, hurting coherence in applications such as cross-document structure construction.

\subsection{Hybrid Pipeline} \label{sec:methods:hybrid}

\subsubsection{Table Generation} \label{sec:methods:hybrid:table}

\begin{table*}[!h]
\small
    \centering
    \renewcommand\arraystretch{1.0}
    \scalebox{0.83}{
    \begin{tabular}{p{4.2cm}R{1cm}P{2.5cm}P{0.7cm}P{1cm}P{0.7cm}P{0.5cm}P{0.5cm}P{0.5cm}L{0.9cm}L{0.9cm}P{0.7cm}}
        \toprule
        \multirow{2}{*}{\textbf{Dataset}}  & \multirow{2}{*}{\textbf{Size}} &  \multirow{2}{*}{\textbf{Domain}} &  \multirow{2}{*}{\textbf{T2S}} & \multirow{2}{*}{\textbf{Annot.}} & \multirow{2}{*}{\textbf{Sch.}} & \multicolumn{3}{c}{\textbf{Text Unit}} & \multicolumn{2}{c}{\textbf{Complexity}}  & \multirow{2}{*}{\textbf{GTS}} \\ 
        \cmidrule(r){7-9} \cmidrule(r){10-11}
        & & & & & & \footnotesize Sent. & \footnotesize Para. & \footnotesize Doc.  & \footnotesize \textbf{Reas.} & \footnotesize \textbf{Struct.} \\ 
        \midrule
        \multicolumn{12}{c}{\textit{Text-Table Datasets}} \\
        \rowcolor[HTML]{DAE8FC}
        WikiBio{\tiny ~\cite{DBLP:conf/emnlp/LebretGA16}} & 728K &  Open-Domain &  \ding{55} & \ding{55} & $\infty$ & \ding{51}& & & \ding{72} & \ding{105} & \ding{55} \\
        E2E{\tiny ~\cite{DBLP:conf/sigdial/NovikovaDR17}} & 50.6K & Restaurant & \ding{55} & \ding{51} & $\square$ & & \ding{51} & & \ding{72} & \ding{105} & \ding{51} \\
        \rowcolor[HTML]{DAE8FC}
        Rotowire{\tiny ~\cite{DBLP:conf/emnlp/WisemanSR17}} & 4.9K &   Sports & \ding{55} &\ding{51} & $\square$ &  & & \ding{51} &  \ding{72}\ding{72} & \ding{105}\ding{105} & \ding{51} \\
        {WikiTableText}{\tiny ~\cite{DBLP:conf/aaai/BaoTDYLZZ18}} & 5.0K &   Open-Domain & \ding{55} &\ding{55} & $\infty$ &  &\ding{51}  & &  \ding{72} & \ding{105} & \ding{51} \\
        \rowcolor[HTML]{DAE8FC}
        {MLB}{\tiny ~\cite{DBLP:conf/acl/PuduppullyDL19}} & 26.3K & Sports & \ding{55} &\ding{51} & $\square$ &  &  \ding{51} & &  \ding{72}\ding{72} & \ding{105}\ding{105} & \ding{51} \\
        {WikiTablePara}{\tiny ~\cite{DBLP:journals/coling/LahaJMS19}} & 171 & Open-Domain & \ding{55} &\ding{51} & $\infty$ &  & \ding{51} & &  \ding{72} & \ding{105}\ding{105} & \ding{51} \\
        \rowcolor[HTML]{DAE8FC}
        WikiTableT{\tiny ~\cite{DBLP:conf/acl/ChenWG21}} & 1.5M & Open-Domain & \ding{55} &\ding{51} & $\infty$ &  & \ding{51} & &  \ding{72} & \ding{105}\ding{105} & \ding{55} \\
        InstructIE{\tiny ~\cite{DBLP:conf/emnlp/Jiao0LZOJ023}} & 14.7K & Open-Domain & \ding{51} & \ding{51} & $\infty$ & & \ding{51} & &\ding{72}\ding{72} &\ding{105}\ding{105} & \ding{51} \\
                \rowcolor[HTML]{DAE8FC}
        {arxivDIGESTables}{\tiny ~\cite{DBLP:conf/emnlp/NewmanLNSFKWCL24}} & 2.2K & arXiv & \ding{51} &\ding{51} & $\infty$ &  & & \ding{51} &  \ding{72}\ding{72}\ding{72} & \ding{105}\ding{105} & \ding{51} \\
        CPL{\tiny~\cite{DBLP:conf/emnlp/JiangLZMCC24}} & 850 & Law\textsuperscript{\tiny \ding{62}} & \ding{51} &\ding{51} & $\infty$ &  & & \ding{51} &  \ding{72}\ding{72} & \ding{105}\ding{105} & \ding{51} \\
                \rowcolor[HTML]{DAE8FC}
        CT-Eval{\tiny~\cite{DBLP:journals/corr/abs-2405-12174}}  & 86.6K & Open-Domain\textsuperscript{\tiny \ding{62}} & \ding{51} &\ding{51} & $\infty$ &  & \ding{51} & &  \ding{72} & \ding{105}\ding{105} & \ding{51} \\
        {DescToTTo}{\tiny~\cite{DBLP:conf/emnlp/RamuGB24}} & 1.3K & Open-Domain & \ding{51} &\ding{51} & $\infty$ &  & \ding{51} &  & \ding{72} & \ding{105}\ding{105}\ding{105} & \ding{51} \\
                \rowcolor[HTML]{DAE8FC}
        {LiveSum}{\tiny~\cite{DBLP:conf/emnlp/DengC00FZYS24}} & 3.8K & Sports & \ding{51} &\ding{51} & $\square$ &  & & \ding{51} &  \ding{72}\ding{72}\ding{72} & \ding{105} & \ding{51} \\
        {Struc-Bench} Table{\tiny ~\cite{DBLP:conf/naacl/TangZPZZCG24}} & 4.1K & Sports & \ding{55} &\ding{51} & $\square$ &  & & \ding{51} &  \ding{72}\ding{72} & \ding{105}\ding{105} & \ding{51} \\
        \midrule 
        \multicolumn{12}{c}{\textit{Text-Graph Datasets}} \\
        \rowcolor[HTML]{DAE8FC}
        {WebNLG}{\tiny~\cite{DBLP:conf/inlg/GardentSNP17}} & 25.3K & Open-Domain & \ding{55} &\ding{51} & $\square$ &  & \ding{51} & &  \ding{72} & \ding{105} & \ding{51} \\
        {SCIERC}{\tiny~\cite{DBLP:conf/emnlp/LuanHOH18}} & 500 & AI Papers & \ding{55} &\ding{51} & $\square$ &  & \ding{51} &   & \ding{72}\ding{72} & \ding{105}\ding{105} & \ding{51} \\
        \rowcolor[HTML]{DAE8FC}
        {AGENDA}{\tiny~\cite{DBLP:conf/naacl/Koncel-Kedziorski19}} & 40K & AI Papers & \ding{55} &\ding{55} & $\square$ &  & \ding{51} & &  \ding{72}\ding{72} & \ding{105}\ding{105} & \ding{55} \\
        {GenWiki}{\tiny~\cite{DBLP:conf/coling/JinGQZ20}} & 1.3M & Open-Domain & \ding{55} &\ding{55} & $\square$ &  & \ding{51} & &  \ding{72} & \ding{105} & \ding{51} \\
        \rowcolor[HTML]{DAE8FC}
        {DART}{\tiny~\cite{DBLP:conf/naacl/NanRZRSHTVVKLIP21}} & 82.2K &Open-Domain & \ding{55} &\ding{51} & $\square$ &  & \ding{51}& &  \ding{72}\ding{72} & \ding{105} & \ding{51} \\
        {EventNarrative}{\tiny~\cite{DBLP:conf/nips/ColasSWW21}} & 224K & EventKG & \ding{55} &\ding{55} & $\square$ &  & \ding{51} & &  \ding{72}\ding{72}\ding{72} & \ding{105}\ding{105}\ding{105} & \ding{55} \\
        \rowcolor[HTML]{DAE8FC}
        {KeLM}{\tiny~\cite{DBLP:conf/naacl/AgarwalGSA21}} & 18M & Open-Domain & \ding{55} &\ding{55} & $\square$ &  \ding{51} & & & \ding{72}\ding{72} & \ding{105} & \ding{55} \\
        {REBEL}{\tiny~\cite{DBLP:conf/emnlp/CabotN21}} & 879K &Open-Domain & \ding{55} &\ding{55} & $\square$ &  \ding{51} &  & &\ding{72}\ding{72} & \ding{105} & \ding{55} \\
        \rowcolor[HTML]{DAE8FC}
        {Text2MindMap}{\tiny~\cite{DBLP:conf/emnlp/HuGZGS21}} & 44.6K & News & \ding{51} & \ding{51} & $\square$ &  &  &  \ding{51} & \ding{72}\ding{72} & \ding{105}\ding{105} & \ding{51} \\
        {InstructIE}{\tiny~\cite{DBLP:journals/corr/abs-2305-11527}} & 362K & Open-Domain\textsuperscript{\tiny \ding{62}} & \ding{51} & \ding{55} & $\square$ & & \ding{51} & & \ding{72}\ding{72} & \ding{105}\ding{105} & \ding{51} \\
        \rowcolor[HTML]{DAE8FC}
        {MINE}{\tiny~\cite{mo2025kggenextractingknowledgegraphs}}  & 100 & Open-Domain & \ding{51} &\ding{55} & $\infty$ & &   & \ding{51} &\ding{72}\ding{72}\ding{72} & \ding{105}\ding{105}\ding{105} & \ding{55} \\    
        \midrule 
        \multicolumn{12}{c}{\textit{Text-Chart Datasets}} \\
        \rowcolor[HTML]{DAE8FC}
        {AutoChart}{\tiny~\cite{DBLP:conf/ranlp/ZhuRLLC21}} & 10.2K & Data Analysis & \ding{55} &\ding{51} & $\square$ & & \ding{51}  & &\ding{72}\ding{72} & \ding{105} & \ding{55} \\
        {Pew}{\tiny~\cite{DBLP:conf/acl/KantharajLLMTHJ22}} & 9.3K & Society \& Policy & \ding{55} &\ding{51} & $\square$ & & \ding{51}  & &\ding{72}\ding{72}\ding{72} & \ding{105}\ding{105}\ding{105} & \ding{51} \\
        \rowcolor[HTML]{DAE8FC}
        {Statista}{\tiny~\cite{DBLP:conf/acl/KantharajLLMTHJ22}} & 34.8K &Market \& Stats & \ding{55} &\ding{51} & $\square$ & &\ding{51}  &  & \ding{72}\ding{72} & \ding{105}\ding{105} & \ding{51} \\
        {Text2Chart}{\tiny~\cite{DBLP:conf/pakdd/RashidJHRZMMS22}} & 717 &Data Analysis & \ding{51} &\ding{51} & $\square$ &  & \ding{51}   & &\ding{72}\ding{72} & \ding{105}\ding{105} & \ding{51} \\
        \rowcolor[HTML]{DAE8FC}
        {ChartX}{\tiny~\cite{DBLP:journals/corr/abs-2402-12185}} & 48K &Open-Domain & \ding{55} &\ding{51} & $\square$ &  & \ding{51}  & &\ding{72}\ding{72}\ding{72} & \ding{105}\ding{105}\ding{105} & \ding{51} \\
        {Text2Chart31}{\tiny~\cite{DBLP:conf/emnlp/ZadehKKK24}} & 11.1K & Data Analysis & \ding{51} &\ding{55} & $\square$ & & \ding{51}   & &\ding{72}\ding{72}\ding{72} & \ding{105}\ding{105}\ding{105} & \ding{51} \\     
        \bottomrule
    \end{tabular}}
    \vspace{-7pt}
    \caption{\small Comparison and classification of existing text-to-structure generation benchmarks across key dimensions. \textbf{Size} denotes the number of (text, data) pairs in the dataset, including the training set. \textbf{Domain} shows the text source domain. \textbf{T2S} (Text-to-Structure) specifies whether the dataset is specifically designed for text-to-structure tasks. \textbf{Annot.} (Annotation) indicates human annotation involvement. \textbf{Sch.} (Schema) shows schema limitedness (`$\square$' for limited, `$\infty$' for unlimited). \textbf{Text Unit} shows the granularity level of text (sentence / paragraph / document). \textbf{Reas.} (Reasoning) indicates the complexity of reasoning required for text-to-structure conversion, and \textbf{Struct.} (Structure) shows the structural complexity of the data, where a higher number of `\ding{72}' or `\ding{105}' indicates greater complexity in the corresponding dimension. \textbf{GTS} (Gold Test Set) denotes manual verification of the test set. `\ding{62}' indicates the presence of Chinese text in the dataset. See Appendix~\ref{sec:app_dataset_cat} for more details.} 
    \label{tab:dataset}
    \vspace{-7pt}
\end{table*}

Harnessing table's natural triple structure (row, column, cell)~\cite{DBLP:conf/emnlp/DengC00FZYS24,DBLP:journals/corr/abs-2403-14457}, it is common to let LLMs perform either tuple extraction from text~\cite{DBLP:journals/corr/abs-2304-08085}, or decomposition of text into atomic propositions~\cite{DBLP:conf/emnlp/HoyleSGR23}, which subsequently serves as input for three approaches: \emph{Code Aggregation}~\cite{DBLP:journals/corr/abs-2308-12950,DBLP:conf/emnlp/DengC00FZYS24} exploits LLM's coding versatility for efficient data integration but falters on complicated structures and scalability. \emph{KG with RAG} builds knowledge graphs for noise-robust RAG-based generation at the cost of flexibility and scalability~\cite{DBLP:conf/emnlp/JiangLZMCC24}. \emph{Schema-Based Inference} models both qualitative nuances and quantitative facts despite similar scalability limitations\cite{DBLP:conf/acl/AhujaBBG25}.

\subsubsection{Graph Generation} \label{sec:methods:hybrid:graph}

KG construction begins with Named Entity Recognition (NER), Relation Extraction (RE), and disambiguation~\cite{DBLP:journals/corr/abs-1909-06058,DBLP:conf/emnlp/HeC21,DBLP:journals/corr/abs-2305-11527}, but is susceptible to noise and hallucinations. Therefore, structure-based denoising aids reliability~\cite{DBLP:conf/aaai/Xie0LL18,DBLP:conf/emnlp/DengW0LS23,DBLP:journals/debu/Bai0Z0FHDCZTGXL25}, and LLM-guided verification boosts accuracy~\cite{DBLP:conf/emnlp/0010WWS0H23,huang2024llmsgoodgraphjudger}. Clustering-based node merging alleviates sparsity but threatens semantic oversimplification~\cite{mo2025kggenextractingknowledgegraphs}. Mindmaps~\cite{DBLP:conf/acl/JainMP24}, on the other hand, take a hierarchical approach: rooted iterative prompting guarantees structural clearness but at the expense of the dense interlinking that is characteristic of KGs.

\subsubsection{Charts Generation}\label{sec:methods:hybrid:chart}

Text-to-chart generation generally employs a two-phase pipeline: \emph{Data point identification}, which has inherent trade-offs: table-based extraction is structured but inflexible~\cite{DBLP:conf/pakdd/RashidJHRZMMS22}, while text-based approaches are more adaptable but may be inconsistent~\cite{DBLP:journals/corr/abs-2410-14331}; \emph{Vision encoding}, which suffers compounded constraints: unclear data hinders chart-type prediction, while visualization synthesis amplifies earlier mistakes~\cite{DBLP:conf/pakdd/RashidJHRZMMS22,DBLP:journals/corr/abs-2410-14331,DBLP:conf/emnlp/ZadehKKK24}. Both phases necessitate more resilient approaches.

\section{Datasets} \label{sec:datasets}

Table~\ref{tab:dataset} provides a comprehensive list of all currently available benchmarks and contrasts them along eight axes. Most structure-to-text datasets can be adapted for text-to-structure tasks~\cite{DBLP:conf/emnlp/WisemanSR17,DBLP:conf/inlg/ObeidH20,DBLP:conf/ranlp/ZhuRLLC21,DBLP:conf/acl/KantharajLLMTHJ22,DBLP:conf/emnlp/0003ZY23,DBLP:journals/tkde/LinRLW24}, although information loss sometimes hinders suitability~\cite{DBLP:conf/emnlp/NieWYPL18,DBLP:conf/acl/ChenCSCW20,DBLP:conf/emnlp/ParikhWGFDYD20}, prompting dedicated text-to-structure dataset creation. Datasets are categorized by human supervision, schema constraints, text length, and exclusive human-annotated test sets. Complexity is quantified via structural complexity of outputs and reasoning complexity in generation (see Appendix~\ref{sec:app_datasets} for further details). Analysis reveals significant gaps: \textit{Text-Table} datasets dominate, yet few demand high complexity; \textit{Text-Graph} datasets generally lack crucial human supervision; and \textit{Text-Chart} datasets are scarce.  Future work should prioritize annotated, high-complexity datasets, especially for underrepresented graph and chart domains, to ensure balanced progress.

\label{sec:datasets:existing}

\section{Evaluation Metrics} \label{sec:metrics}

\begin{figure}[!t]
    \centering
    \includegraphics[width=\linewidth]{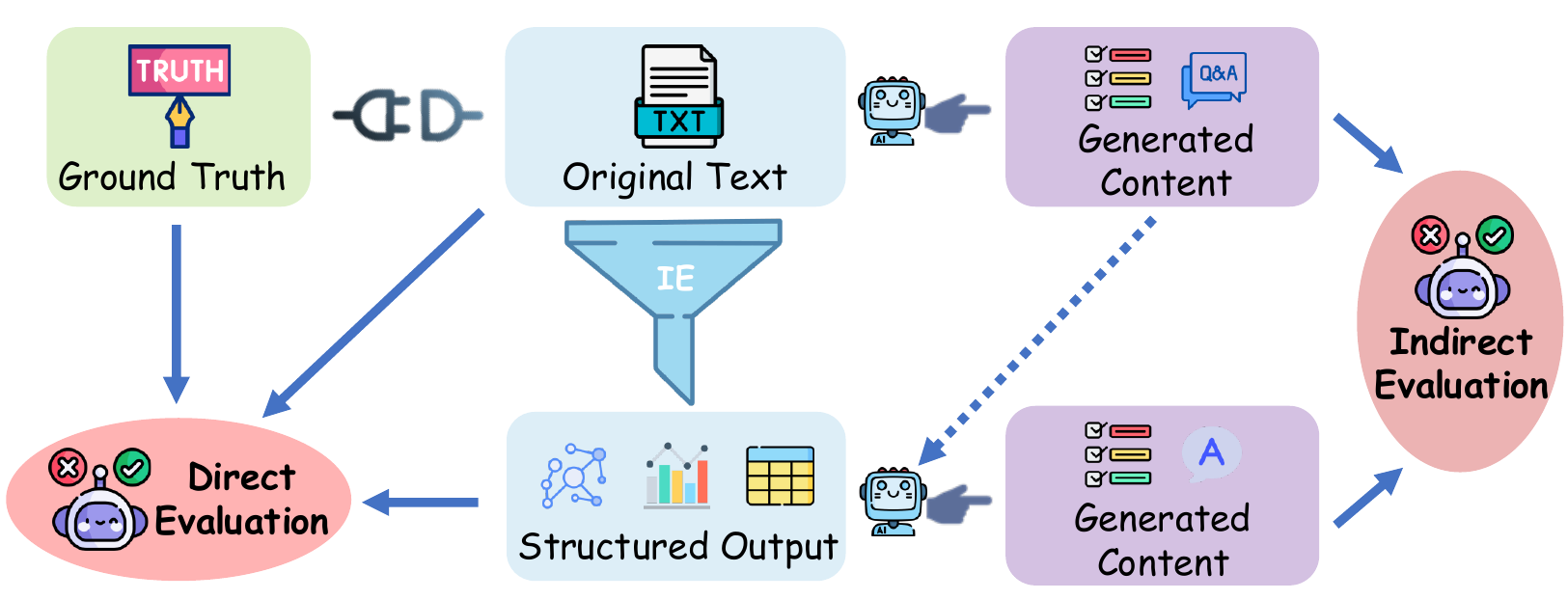}
    \caption{Comparison of direct evaluation and indirect evaluation methods for structured outputs.}
    \vspace{-7pt}
    \label{fig:cmp_eval}
    \vspace{-10pt}
\end{figure}

Evaluating complex structured outputs poses multifaceted challenges distinct from traditional IE tasks. While the latter often benefit from well-defined formats and objective standards~\cite{DBLP:journals/fcsc/XuCPZXZWZWC24}, the inherent diversity and complexity of structured outputs necessitate a more comprehensive evaluation framework.
As shown in Figure~\ref{fig:cmp_eval}, after generating \textit{structured output} from the \textit{original text}, prevailing methods bifurcate into: (1) \textbf{\textit{Direct Evaluation}} jointly assesses the \textit{ground truth}, \textit{original text}, and \textit{structured output}~(\S\ref{sec:metrics:human}, \S\ref{sec:metrics:rule}, \S\ref{sec:metrics:llm}); (2) \textbf{\textit{Indirect Evaluation}} first generates intermediate \textit{content} (e.g., propositions, qustion-answer pairs) from the \textit{original text}, then uses this \textit{content} and the \textit{structured output} to produce further \textit{content} (e.g., answers), and finally compares the two \textit{generated content} sets~(\S\ref{sec:metrics:llm}). Through analysis of this established framework, we identify limitations that motivate our novel evaluation paradigm.

\subsection{Human-based Evaluation} \label{sec:metrics:human}
Human-based evaluation remains the gold standard, assessing content coverage, structural validity, and factual consistency~\cite{DBLP:conf/emnlp/Jiao0LZOJ023,DBLP:conf/naacl/TangZPZZCG24,DBLP:conf/acl/JainMP24}. It can also be evaluated by manual pairwise comparison~\cite{DBLP:conf/emnlp/ZadehKKK24}, processing pipeline stages~\cite{DBLP:conf/pakdd/RashidJHRZMMS22,DBLP:conf/emnlp/DengC00FZYS24}, or indirectly via comprehension tasks~\cite{DBLP:conf/acl/JainMP24}. While human evaluation provides accurate results, it is costly in terms of time and resources, highlighting the complexity of quality that automated metrics must capture.

\subsection{Rule-based Evaluation} \label{sec:metrics:rule}
Metrics like Exact Match, ROUGE-L~\cite{lin-2004-rouge}, Levenshtein Distance~\cite{DBLP:journals/corr/abs-1101-1232}, and chrF~\cite{DBLP:conf/wmt/Popovic15} rely on predefined rules focusing on surface-level features: token overlap, sequence edits, or n-gram alignment~\cite{DBLP:conf/wmt/Post18,9649217,DBLP:conf/acl/0001ZL22,DBLP:conf/naacl/TangZPZZCG24}. 
Although ROUGE-L shows the strongest correlation with human evaluation among these metrics~\cite{DBLP:conf/acl/LiWSZWS23,DBLP:conf/nips/WangIDHKCWMSBH23}, it still fails to capture semantic and structural aspects, leading to misalignment in structured output assessment despite its convenience.

\subsection{Model-based Evaluation}\label{sec:metrics:llm}

\paragraph{Direct Scoring}
Learning-based methods use pre-trained models like SentenceBERT~\cite{DBLP:conf/emnlp/ReimersG19}, BERTScore~\cite{DBLP:conf/iclr/ZhangKWWA20}, BLEURT~\cite{DBLP:conf/acl/SellamDP20}, and BARTScore~\cite{DBLP:conf/nips/YuanNL21} to evaluate text by semantic and contextual similarity, demonstrating better alignment with human judgements than token-matching rules on semantics~\cite{DBLP:conf/emnlp/Jiao0LZOJ023,DBLP:conf/naacl/TangZPZZCG24,DBLP:conf/emnlp/DengC00FZYS24}. However, these models remain inadequate for assessing structural correctness, as they lack explicit schema understanding and relational reasoning, and are insensitive to fine-grained content variations critical for quality assessment. On the other hand, direct prompting of LLMs partially addresses this gap by evaluating content-structure similarity via Chain-of-Thought prompting~\cite{DBLP:conf/acl/ChiangL23,DBLP:conf/emnlp/LiuIXWXZ23,DBLP:conf/naacl/TangZPZZCG24} although its reliability depends on prompt design and model biases.

\paragraph{Indirect Scoring}
Complex structured outputs often lack unique ground truth due to valid variations, like row/column order, equivalent values, or complete absence~\cite{DBLP:conf/lrec/GuoDVA20,DBLP:conf/acl/LiWSZWS23,DBLP:conf/naacl/TangZPZZCG24,DBLP:conf/acl/JainMP24}, necessitating automated quality assessment methods: NLI-based alignment provides superior robustness to structural diversity than rigid rule-matching~\cite{DBLP:journals/corr/abs-1907-11692,DBLP:conf/emnlp/RamuGB24}; QA-based evaluation~\cite{DBLP:conf/acl/JainMP24,DBLP:conf/emnlp/DengC00FZYS24} better assesses how well the structured outputs retain original information. Although these indirect evaluation approaches are more robust and generalizable, they demand stronger reasoning capabilities from models, such as the techniques from TableQA~\cite{DBLP:journals/pvldb/ZhangHFCDP24,DBLP:conf/iclr/0002ZLEP0MFSLP24}, KGQA~\cite{DBLP:conf/emnlp/JiangZDYZW23,DBLP:conf/iclr/LuoLHP24,DBLP:conf/acl/JinXZRZL0TWM024}, and chart reasoning~\cite{DBLP:conf/emnlp/MasryKLHJ23,DBLP:conf/eacl/AkhtarCS23,DBLP:conf/acl/MasrySPHJ24,DBLP:conf/emnlp/ZhangHXYXJZ024} to avoid error propagation while assessing outputs.

\section{{Text-to-Structure} (T2S) Benchmark} \label{sec:advie}

To address the limitations of existing evaluation frameworks summarized in Section~\ref{sec:metrics}, we propose a novel framework capable of accurately and comprehensively evaluating structured outputs. Qualitatively, we address the previous vague criteria by implementing concrete metrics: an F1-style semantic metric using LLM-as-judge for \textbf{content} evaluation, overcoming limitations of string-based metrics like ROUGE and BERTScore~\cite{DBLP:conf/naacl/TangZPZZCG24}, and defining two new orthogonal dimensions for \textbf{structure}, \textit{alignment} for format consistency and \textit{clarity} for unambiguous representations, inspired by header evaluation of prior works~\cite{DBLP:conf/acl/0001ZL22,DBLP:conf/emnlp/Jiao0LZOJ023} and ~\citet{DBLP:conf/acl/JainMP24}. To validate our framework's superiority, we introduce a novel Text-to-Structure (T2S) benchmark, detailing: criteria (\S\ref{sec:score_criteria}), datasets and models (\S\ref{sec:dataandmodel}), results (\S\ref{sec:results}), and validation of effectiveness (\S\ref{sec:eval}).

\subsection{Scoring Criteria} \label{sec:score_criteria}

\begin{table*}[t]
    \centering
    \scalebox{0.70}{
    \begin{tabular}{p{3.6cm}|P{1.0cm}P{1.0cm}|P{1.0cm}P{1.0cm}|P{1.0cm}P{1.0cm}|P{1.0cm}P{1.0cm}|P{1.0cm}P{1.0cm}|P{1.0cm}P{1.0cm}}
        \toprule
        \multirow{4}{*}{\textbf{Model}}  & \multicolumn{6}{c|}{\textit{Table}}  & \multicolumn{4}{c|}{\textit{Graph}} & \multicolumn{2}{c}{\textit{Chart}} \\
        \cmidrule(r){2-7}  \cmidrule(r){8-11} \cmidrule(r){12-13}
        & \multicolumn{2}{c}{\textbf{InstructIE}} &  \multicolumn{2}{c}{\textbf{Struc-Bench}} & \multicolumn{2}{c|}{\textbf{LiveSum}} & \multicolumn{2}{c}{\textbf{DART}} & \multicolumn{2}{c|}{\textbf{InstructIE}} & \multicolumn{2}{c}{\textbf{Text2Chart31}} \\
        & \multicolumn{2}{c}{{\footnotesize \cite{DBLP:conf/emnlp/Jiao0LZOJ023}}} &  \multicolumn{2}{c}{\footnotesize \cite{DBLP:conf/naacl/TangZPZZCG24}} & \multicolumn{2}{c|}{\footnotesize \cite{DBLP:conf/emnlp/DengC00FZYS24}} & \multicolumn{2}{c}{\footnotesize \cite{DBLP:conf/naacl/NanRZRSHTVVKLIP21}} & \multicolumn{2}{c|}{\footnotesize \cite{DBLP:journals/corr/abs-2305-11527}} & \multicolumn{2}{c}{\footnotesize \cite{DBLP:conf/emnlp/ZadehKKK24}} \\
         \cmidrule(r){2-3} \cmidrule(r){4-5} \cmidrule(r){6-7} \cmidrule(r){8-9} \cmidrule(r){10-11} \cmidrule(r){12-13}
        & \textsc{Faith}. & \textsc{Coh.} & \textsc{Faith}. & \textsc{Coh.}   & \textsc{Faith}. & \textsc{Coh.}   & \textsc{Faith}. & \textsc{Coh.}   & \textsc{Faith}. & \textsc{Coh.}   & \textsc{Faith}. & \textsc{Coh.}    \\
        \midrule
        \multicolumn{13}{l}{$\vardiamondsuit$ \emph{\textbf{SFT}}} \\
Mistral-7B  & \textcolor{below60}{58.08} & \textcolor{above60}{68.85}  & \textcolor{above80}{87.65} & \textcolor{above90}{90.05}  & \textcolor{below60}{53.31} & \textcolor{above90}{98.54}  & \textcolor{above70}{75.07} & \textbf{\textcolor{above80}{83.17}}  & \textcolor{below60}{44.58} & \textcolor{above60}{69.67}  & \textcolor{above60}{64.54} & \textcolor{above60}{63.41} \\
Llama-3.1-8B  & \textcolor{below60}{59.31} & \textcolor{above60}{66.44}  & \textcolor{above80}{85.01} & \textcolor{above80}{85.95}  & \textcolor{above60}{61.34} & \textcolor{above90}{90.30}  & \textcolor{below60}{38.94} & \textcolor{below60}{55.13}  & \textcolor{below60}{34.93} & \textcolor{above60}{64.99}  & \textcolor{above70}{77.06} & \textcolor{above70}{77.25} \\
R1-Distill-Llama-8B  & \textcolor{above60}{62.86} & \textcolor{above60}{68.08}  & \textcolor{below60}{54.56} & \textcolor{below60}{58.97}  & \textcolor{below60}{43.27} & \textcolor{above80}{86.22}  & \textcolor{below60}{59.03} & \textcolor{above60}{69.43}  & \textcolor{below60}{37.27} & \textcolor{above60}{69.73}  & \textcolor{below60}{41.66} & \textcolor{below60}{40.45} \\
        \midrule
        \multicolumn{13}{l}{$\vardiamondsuit$ \emph{\textbf{Zero-Shot}}} \\
Phi-3-medium  & \textcolor{below60}{29.15} & \textcolor{below60}{39.89}  & \textcolor{above70}{74.04} & \textcolor{above80}{83.99}  & \textcolor{below60}{41.61} & \textcolor{above90}{91.73}  & \textcolor{above60}{64.60} & \textcolor{above70}{72.59}  & \textcolor{below60}{43.40} & \textcolor{above70}{71.44}  & \textcolor{above60}{60.24} & \textcolor{above60}{60.45} \\
Phi-3-medium~(\textsc{CoT})  & \textcolor{below60}{30.36} & \textcolor{below60}{37.23}  & \textcolor{above70}{73.42} & \textcolor{above80}{81.06}  & \textcolor{below60}{46.36} & \textcolor{above90}{93.19}  & \textcolor{above60}{64.88} & \textcolor{above70}{72.32}  & \textcolor{below60}{41.28} & \textcolor{above70}{70.44}  & \textcolor{below60}{57.75} & \textcolor{below60}{58.27} \\
Mistral-7B  & \textcolor{below60}{53.00} & \textcolor{below60}{59.73}  & \textcolor{below60}{56.19} & \textcolor{above60}{67.05}  & \textcolor{below60}{30.14} & \textcolor{above80}{87.26}  & \textcolor{above70}{70.27} & \textcolor{above70}{73.06}  & \textcolor{below60}{42.74} & \textcolor{above60}{66.54}  & \textcolor{below60}{52.25} & \textcolor{below60}{52.03} \\
Mistral-7B~(\textsc{CoT})  & \textcolor{below60}{54.02} & \textcolor{above60}{60.37}  & \textcolor{below60}{56.25} & \textcolor{above70}{70.34}  & \textcolor{below60}{41.27} & \textcolor{above80}{85.97}  & \textcolor{above60}{64.68} & \textcolor{above70}{70.96}  & \textcolor{below60}{33.35} & \textcolor{above60}{66.01}  & \textcolor{below60}{45.13} & \textcolor{below60}{46.51} \\
Mixtral-8x22B  & \textcolor{below60}{56.35} & \textcolor{above60}{68.25}  & \textcolor{above80}{84.16} & \textcolor{above80}{85.20}  & \textcolor{below60}{49.12} & \textcolor{above80}{87.91}  & \textcolor{above60}{65.70} & \textcolor{above70}{71.07}  & \textcolor{below60}{40.07} & \textcolor{above60}{68.89}  & \textcolor{above60}{61.62} & \textcolor{above60}{62.15} \\
Mixtral-8x22B~(\textsc{CoT})  & \textcolor{below60}{58.27} & \textcolor{above60}{68.29}  & \textcolor{above80}{80.55} & \textcolor{above70}{74.94}  & \textcolor{below60}{54.87} & \textcolor{above80}{87.37}  & \textcolor{above70}{71.27} & \textcolor{above70}{74.53}  & \textcolor{below60}{47.47} & \textcolor{above70}{71.29}  & \textcolor{below60}{58.92} & \textcolor{below60}{58.89} \\
Llama-3.1-405B  & \textcolor{above60}{63.88} & \textcolor{above60}{69.63}  & \textbf{\textcolor{above80}{89.43}} & \textcolor{above90}{90.43}  & \textcolor{above60}{62.29} & \textcolor{above90}{99.11}  & \textcolor{above60}{68.74} & \textcolor{above70}{75.61}  & \textcolor{below60}{46.72} & \textcolor{above70}{75.82}  & \textcolor{above80}{86.49} & \textcolor{above80}{86.62} \\
Llama-3.1-405B~(\textsc{CoT})  & \textcolor{above60}{64.54} & \textcolor{above70}{72.80}  & \textcolor{above80}{89.38} & \textcolor{above90}{90.74}  & \textcolor{above60}{66.25} & \textcolor{above90}{98.64}  & \textcolor{above70}{72.31} & \textcolor{above70}{78.09}  & \textcolor{below60}{47.94} & \textcolor{above70}{78.48}  & \textcolor{above80}{87.23} & \textcolor{above80}{86.40} \\
Llama-3.3-70B  & \textcolor{above60}{62.88} & \textcolor{above60}{69.19}  & \textcolor{above80}{80.58} & \textcolor{above70}{80.00}  & \textcolor{below60}{58.48} & \textbf{\textcolor{above90}{99.19}}  & \textcolor{above70}{71.09} & \textcolor{above70}{77.54}  & \textcolor{below60}{47.11} & \textcolor{above70}{76.30}  & \textcolor{above80}{87.44} & \textcolor{above80}{87.12} \\
Llama-3.3-70B~(\textsc{CoT})  & \textcolor{above60}{65.34} & \textcolor{above60}{69.46}  & \textcolor{above80}{86.47} & \textcolor{above80}{85.84}  & \textcolor{above60}{62.79} & \textcolor{above90}{98.68}  & \textcolor{above70}{77.11} & \textcolor{above70}{78.99}  & \textcolor{below60}{46.56} & \textcolor{above70}{78.51}  & \textbf{\textcolor{above80}{88.78}} & {\textcolor{above80}{87.83}} \\
GPT-3.5-turbo  & \textcolor{below60}{55.09} & \textcolor{above60}{69.01}  & \textcolor{above70}{78.44} & \textcolor{above80}{85.37}  & \textcolor{below60}{45.38} & \textcolor{above70}{79.18}  & \textcolor{above60}{69.75} & \textcolor{above70}{76.72}  & \textcolor{below60}{41.89} & \textcolor{above70}{73.34}  & \textcolor{above80}{85.17} & \textcolor{above80}{84.79} \\
GPT-3.5-turbo~(\textsc{CoT})  & \textcolor{below60}{58.44} & \textcolor{above60}{68.80}  & \textcolor{above70}{71.88} & \textcolor{above80}{81.54}  & \textcolor{below60}{45.09} & \textcolor{above80}{86.37}  & \textcolor{above70}{70.17} & \textcolor{above70}{75.99}  & \textcolor{below60}{41.78} & \textcolor{above70}{73.51}  & \textcolor{above80}{83.06} & \textcolor{above80}{83.62} \\
GPT-4o  & \textcolor{above70}{70.72} & \textbf{\textcolor{above70}{76.36}}  & \textcolor{above80}{87.35} & {\textcolor{above90}{92.22}}  & \textcolor{above60}{63.21} & \textcolor{above90}{91.99}  & \textcolor{above70}{74.15} & \textcolor{above70}{77.18}  & \textcolor{below60}{46.77} & \textcolor{above70}{77.00}  & \textcolor{above80}{85.48} & \textcolor{above80}{86.01} \\
GPT-4o~(\textsc{CoT})  & \textcolor{above60}{68.11} & \textcolor{above70}{74.40}  & \textcolor{above80}{86.05} & \textcolor{above90}{90.03}  & \textcolor{above60}{62.16} & \textcolor{above90}{98.50}  & \textbf{\textcolor{above70}{79.89}} & \textcolor{above70}{79.33}  & {\textcolor{below60}{53.28}} & \textcolor{above70}{77.65}  & \textcolor{above80}{85.99} & \textcolor{above80}{86.63} \\
QwQ-32B  & \textcolor{above60}{{68.23}} & \textcolor{above60}{{68.79}} & \textcolor{above80}{{81.21}} & \textcolor{above80}{{87.12}} & \textcolor{below60}{{58.06}} & \textcolor{above90}{{96.20}} & \textcolor{above60}{{69.42}} & \textcolor{above70}{{78.02}} & \textcolor{below60}{{56.30}} & \textcolor{above80}{{82.15}} & \textcolor{above80}{{86.62}} & \textcolor{above80}{{86.71}} \\
DeepSeek-R1  & \textcolor{above70}{{\textbf{76.86}}} & \textcolor{above70}{{74.75}} & \textcolor{above80}{{87.77}} & \textcolor{above90}{{91.05}} & \textcolor{above60}{{64.89}} & \textcolor{above90}{{96.85}} & \textcolor{above70}{{78.25}} & \textcolor{above70}{{79.89}} & \textcolor{below60}{{52.15}} & \textcolor{above70}{{79.23}} & \textcolor{above80}{{86.07}} & \textcolor{above80}{{86.90}} \\
Grok-3-mini & \textcolor{above70}{{75.57}} & \textcolor{above70}{{75.40}} & \textcolor{above80}{{85.60}} & \textcolor{above90}{{\textbf{92.28}}} & \textcolor{above60}{{66.35}} & \textcolor{above90}{{95.26}} & \textcolor{above70}{{73.32}} & \textcolor{above80}{{80.34}} & \textcolor{below60}{{56.03}} & \textcolor{above80}{{\textbf{82.68}}} & \textcolor{above80}{{87.63}} & \textcolor{above80}{{\textbf{88.65}}} \\
O4-mini & \textcolor{above70}{{75.71}} & \textcolor{above70}{{74.85}} & \textcolor{above80}{{83.45}} & \textcolor{above90}{{90.02}} & \textcolor{above80}{{\textbf{84.48}}} & \textcolor{above90}{{99.16}} & \textcolor{above70}{{74.82}} & \textcolor{above80}{{80.19}} & \textcolor{below60}{{\textbf{56.46}}} & \textcolor{above80}{{82.32}} & \textcolor{above80}{{85.21}} & \textcolor{above80}{{85.93}} \\
\bottomrule
    \end{tabular}}
    \vspace{-7pt}
    \caption{The performance of major LLMs across multiple datasets. The highest results are \textbf{bolded}, with numerical scores categorized into five color-coded intervals: \textcolor{above90}{{[90, 100] - excellent}}, \textcolor{above80}{{[80, 90) - strong}}, \textcolor{above70}{{[70, 80) - moderate}}, \textcolor{above60}{{[60, 70) - limited}}, and \textcolor{below60}{{[0, 60) - insufficient}}, using distinct chromatic gradients for visual differentiation.} 
    \label{tab:rrr}
    \vspace{-10pt}
\end{table*}

We evaluate structured outputs across two dimensions: \textit{content faithfulness} and \textit{structure coherence}. 
\paragraph{Content Faithfulness}
Measures the accuracy and completeness of the generated information, ensuring it reflects the source text without errors or omissions. It uses an F1-style metric combining: (1) \textit{\textbf{precision}}, which identifies conflicts between output and reference data (e.g., factual errors or semantic mismatches) and (2) \textit{\textbf{recall}}, which measures how thoroughly the output captures key details from the source. These are combined into a single score via harmonic mean:
\begin{equation*}
\small
    \textsc{Faithfulness}=2 \cdot \frac{\textit{precision}\cdot \textit{recall}}{\textit{precision}+\textit{recall}}
\end{equation*}
\paragraph{Structure Coherence}
Evaluates logical organization and interoperability of the output. It examines (1) \textit{\textbf{alignment}} with expected formats (e.g., schema consistency, data types) and (2) \textit{\textbf{clarity}} of presentation (e.g., unambiguous representation, redundancy-free content). Again, we employ the harmonic mean for these two orthogonal metrics, as it penalizes imbalance and more accurately reflects joint effectiveness:
\begin{equation*}
\small
    \textsc{Coherence}=2 \cdot \frac{\textit{alignment}\cdot \textit{clarity}}{\textit{alignment}+\textit{clarity}}
\end{equation*}
When applied to agentic workflows, this scoring framework optimizes the performance of advanced IE: \textit{precision} and \textit{recall} validate extracted data fidelity, ensuring agents develop an accurate and comprehensive understanding of data, thereby facilitating the generation of reliable and holistic decisions; while \textit{alignment} and \textit{clarity} govern integration with downstream modules, enhancing interoperability within agentic AI systems and optimizing the efficiency of data interpretation and action. We employ LLM-as-judge with detailed scoring rubrics (0-100 scale) for all four sub-metrics: \textit{precision}, \textit{recall}, \textit{alignment}, and \textit{clarity}. Detailed explanations and implementation are in Appendix~\ref{sec:app_criteria_detail}.

\subsection{Datasets and Models} \label{sec:dataandmodel}
We carefully select six datasets from the openly available sources listed in Table~\ref{tab:dataset}, aiming to encompass a diverse range of complexities, cover multiple languages, and prioritize open-domain datasets. Specifically, three datasets for text-to-table generation: InstructIE, Struc-Bench, and LiveSum; two for text-to-graph generation: DART and InstructIE; and Text2Chart31 for text-to-chart generation. We select several mainstream LLMs for evaluation, which can be divided into two groups: non-thinking models (e.g., Llama, GPT) and thinking models (e.g., R1, O4-mini). For the former group, we also evaluate multiple models under SFT. Other models are evaluated in a zero-shot setting. For non-thinking models, we perform inference at temperature 0; for thinking models, we use the documented recommended hyperparameters, averaging results across five runs. When employing DeepSeek-R1 and GPT-4o as LLM-as-judge, we fix the temperature parameter at 0 to ensure that every evaluation remains deterministic. Further details are provided in Appendix~\ref{sec:app_dataset_selection} and~\ref{sec:app_select_models}.

\subsection{Results} \label{sec:results}

In the Text-to-Table task, LiveSum emerges as the most challenging benchmark in terms of content faithfulness, with all evaluated models (except O4-mini) scoring no higher than 70; InstructIE is intermediate, while Struc-Bench is the simplest, achieving nearly 90. Regarding structure coherence, both LiveSum and Struc-Bench achieve higher scores due to their predefined headers, reflecting strong instruction-following capabilities. In contrast, the absence of fixed headers in InstructIE prompts models to generate more diverse table formats, resulting in lower coherence scores.
In the Text-to-Graph task, the DART dataset demonstrates higher faithfulness compared to InstructIE because both datasets, originally designed for KG-to-text, contain triples often irretrievable from text and this limitation is more pronounced in InstructIE and leads to lower recall. Meanwhile, both datasets show moderate performance in structure coherence, primarily due to challenges in entity and relation naming.
In the Text-to-Chart task, larger models generally exhibit superior performance, consistently achieving scores above 80, while smaller models trained with SFT lag significantly behind.
For further analysis, see Appendix~\ref{sec:app_result_analysis}.

\subsection{Evaluation of Scoring Criteria} \label{sec:eval}

To demonstrate how our proposed scoring criteria address evaluation limitations, we aim to answer the following questions:
\begin{center}
\vspace{-10pt}
\setlength{\tabcolsep}{0pt} 
\begin{tabular}{p{1.1cm}p{6.7cm}}
\textbf{RQ1.} & Do traditional metrics align with human judgment criteria? (\S~\ref{subsubsec:rq1}) \\
\textbf{RQ2.} & Can LLM-based evaluation ensure consistent and nuanced assessment? (\S~\ref{subsubsec:rq2})\\
\end{tabular}  
\end{center}
Furthermore, we provide a case study analyzing traditional metric failures and our framework's strengths~(\S \ref{sec:cass}).

\subsubsection{Validity Gap (RQ1)} \label{subsubsec:rq1}
\begin{table}[!t]
    \centering
    \scalebox{0.68}{
    \begin{tabular}{p{1cm}p{2cm}P{0.7cm}P{0.7cm}P{0.7cm}P{0.7cm}P{0.7cm}P{0.7cm}}
    \toprule
         \multirow{2}{*}{\textbf{Task}} &  \multirow{2}{*}{\textbf{Metric}} &  \multicolumn{3}{c}{\textbf{Faithfulness}} &  \multicolumn{3}{c}{\textbf{Coherence}} \\
         \cmidrule(r){3-5} \cmidrule(r){6-8}
        & &  P($r$) &  S($\rho$) & K($\tau$) & P($r$) &  S($\rho$) & K($\tau$) \\
    \midrule
     \multirow{2}{*}{\textit{Table}}   & ROUGE-L &  0.267 & 0.227 & 0.166 & \textbf{0.607} & \textbf{0.617} & 0.452 \\ 
     & BERTScore & 0.238 & 0.198 & 0.142 & \textbf{0.562} & \textbf{0.576} & 0.415 \\
    \midrule
     \multirow{2}{*}{\textit{Graph}}   & ROUGE-L &  0.237 & 0.118 & 0.093 & 0.318 & 0.165 & 0.127 \\ 
     & BERTScore & 0.100 & 0.043 & 0.036 & 0.211 & 0.105 & 0.078 \\
    \midrule
     \multirow{2}{*}{\textit{Chart}}   & CodeBLEU &  0.249 & 0.186 & 0.144 & 0.266 & 0.297 & 0.220 \\ 
     & METEOR & 0.281 & 0.241 & 0.186 & 0.317 & 0.376 & 0.274 \\
    \bottomrule
    \end{tabular}
    }
    \vspace{-7pt}
    \caption{Correlation coefficients (Pearson/Spearman/Kendall) between traditional metrics and human evaluation across two dimensions on three different tasks.}
    \label{tab:correlation_coefficient_tradition}
    \vspace{-15pt}
\end{table}
For table generation and graph generation tasks, prior works predominantly adopt ROUGE-L and BERTScore by extracting textual content from structured outputs and comparing them with ground truth references~\cite{DBLP:conf/acl/0001ZL22,DBLP:conf/emnlp/Jiao0LZOJ023,huang2024llmsgoodgraphjudger}. For chart generation tasks, existing approaches typically evaluate code similarity between generated and reference codes using metrics like METEOR and CodeBLEU~\cite{DBLP:conf/emnlp/ZadehKKK24}. To investigate whether these metrics align with human evaluation, we employ three annotators to score outputs based on our two orthogonal criteria, \textit{Faithfulness} and \textit{Coherence}, with final scores averaged across annotators. We conduct correlation analyses between each automated metric and human-rated \textit{Faithfulness} and \textit{Coherence} following established methodologies~\cite{DBLP:conf/emnlp/Jiao0LZOJ023}, with results summarized in Table~\ref{tab:correlation_coefficient_tradition}. The table reveals that only in table generation tasks do ROUGE-L and BERTScore exhibit moderate correlations with Coherence ($r,\rho>0.5$), whereas all other metrics exhibit weak alignment with human evaluation. This demonstrates traditional metrics' limitations and underscores the urgent need for specialized metrics capable of reliable quality assessment in this domain.

\subsubsection{Reliability and Discriminability (RQ2)} \label{subsubsec:rq2}
To investigate the alignment between LLM-based scoring and human evaluation under our proposed criteria, three annotators score 30 randomly sampled instances per dataset following the rubric outlined in the evaluation prompt. The Root Mean Square Error (RMSE) between human and LLM scores is averaged across three task categories and aggregated in Table~\ref{tab:rmse}, which reveals that DeepSeek-R1 demonstrates superior accuracy compared to GPT-4o, with subsequent case studies confirming that its enhanced reasoning fidelity directly contributes to reduced prediction errors. Given DeepSeek-R1's current lack of multimodal input capabilities, GPT-4o remains our evaluator for charts to maintain assessment consistency.
\begin{table}[!t]
    \centering
    \scalebox{0.72}{
    \begin{tabular}{p{1cm}p{2.6cm}p{0.9cm}p{0.9cm}p{0.9cm}p{0.9cm}}
    \toprule
         \textbf{Task} & \textbf{Model} &  \textbf{Prec.} &  \textbf{Rec.} &  \textbf{Align.} &  \textbf{Clr.}  \\
    \midrule
     \multirow{2}{*}{\textit{Table}}   & GPT-4o &  14.90 & 15.79 & 23.76 & 15.14 \\ 
     & DeepSeek-R1 & \ \ \textbf{9.37}	& \ \ \textbf{8.59}	&	\ \ \textbf{8.32}&	\ \ \textbf{4.55} \\
    \midrule
    \multirow{2}{*}{\textit{Graph}}   & GPT-4o &  14.76 & 11.08 & \ \ 7.16 & \ \ 7.42 \\ 
     & DeepSeek-R1 &	\ \ \textbf{4.91} & \ \ \textbf{8.91} & \ \ \textbf{7.09} & \ \ \textbf{3.08} \\
     \midrule
     \textit{Chart} & GPT-4o & 15.32 & 14.77 & \ \ 6.12 & \ \ 7.91 \\
    \bottomrule
    \end{tabular}
    }
    \vspace{-7pt}
    \caption{Root Mean Squared Error~(RMSE) between human and LLM scores across three task categories.}
    \label{tab:rmse}
    \vspace{-15pt}
\end{table}
 We further analyze the correlation between human ratings and DeepSeek-R1 scores, computing Pearson's $r$, Spearman's $\rho$, and Kendall's $\tau$ coefficients, with results summarized in Table~\ref{tab:correlation_analysis}. All four evaluation metrics exhibit strong positive correlations with human judgments.
\begin{table}[!t]
    \centering
    \scalebox{0.72}{
    \begin{tabular}{p{2.4cm}P{1.5cm}P{1.5cm}P{1.5cm}P{1.5cm}}
    \toprule
    \multirow{2}{*}{\textbf{Metric}} & \multicolumn{2}{c}{\textbf{Faithfulness}} &  \multicolumn{2}{c}{\textbf{Coherence}} \\
    \cmidrule(r){2-3} \cmidrule(r){4-5}
          &   {Precision} &  {Recall} &  {Alignment}&  {Clarity} \\
    \midrule
    Pearson ($r$) & 0.745 & 0.771 & 0.761 & 0.728 \\
    Spearman ($\rho$) & 0.747 & 0.766 & 0.712 & 0.679 \\
    Kendall ($\tau$) & 0.664 & 0.684 & 0.670 & 0.621 \\
    \bottomrule
    \end{tabular}
    }
    \vspace{-7pt}
    \caption{Correlation coefficients (Person / Spearman / Kendall) of human and LLM scores for four evaluation metrics.}
    \label{tab:correlation_analysis}
    \vspace{-15pt}
\end{table}
 To compare our proposed metrics with conventional metrics, we evaluated five models on the InstructIE~\cite{DBLP:conf/emnlp/Jiao0LZOJ023} dataset using both traditional metrics (ROUGE and BERTScore) and LLM-based assessments (\textsc{Faith.} and \textsc{Coh.} scores from GPT-4o and DeepSeek-R1). The results in Table~\ref{tab:perf_instructie} demonstrate that DeepSeek-R1 scores show a high positive correlation with GPT-4o, eliminating concerns about LLM evaluators favoring their own outputs. While GPT-3.5-turbo achieves the second-highest scores in traditional metrics, it ranks last across all LLM-based evaluations, validating our metrics' capability to detect reliability flaws obscured by conventional measurements.
\begin{table}[!t]
    \centering
    \scalebox{0.72}{
    \begin{tabular}{p{2.6cm}|p{0.8cm}p{0.8cm}|p{0.9cm}p{0.9cm}p{0.7cm}p{0.7cm}}
    \toprule
         \textbf{Model} & \footnotesize \textbf{\textsc{Rouge}} &\textbf{\footnotesize \textsc{Bert}.}  & \footnotesize \textbf{\textsc{Faith}\textsubscript{R1}} & \footnotesize \textbf{\textsc{Faith}\textsubscript{4o}} & \footnotesize \textbf{\textsc{Coh}\textsubscript{R1}} & \footnotesize \textbf{\textsc{Coh}\textsubscript{4o}}  \\
    \midrule
    Llama-3.1-405B & 0.561 & 0.742 & 63.88 & 69.53 & 69.63 & 76.66 \\ 
    Llama-3.3-70B&	0.517&	0.738&	62.88&	67.23&	69.19&	75.08 \\
GPT-3.5-turbo&	{\underline{0.586}}&	{\underline{0.770}}&	55.09&	62.86&	69.01&	75.82 \\
GPT-4o&	\textbf{0.613}&	\textbf{0.772}&	\underline{70.72}&	\underline{73.07}&	\textbf{76.36}&	\textbf{81.44} \\ 
DeepSeek-R1&	0.541&	0.738&	\textbf{76.86}&	\textbf{77.02}&	\underline{74.75}&	\underline{79.81} \\

    \bottomrule
    \end{tabular}
    }
    \vspace{-7pt}
    \caption{Performance comparison of five LLMs on InstructIE~\cite{DBLP:conf/emnlp/Jiao0LZOJ023} dataset through conventional metrics and LLM-based evaluation, and we \textbf{bold} the best performance for each metric.}
    \label{tab:perf_instructie}
    \vspace{-15pt}
\end{table}

\subsubsection{Case Study} \label{sec:cass}
We provide five carefully examined case studies tackling important shortcomings in automatic evaluation metrics in Appendix~\ref{app:case_study}. Our examination uncovers systematic misalignment situations where conventional metrics (BERTScore, ROUGE, etc.) incorrectly prefer outputs with: structural incompleteness (Case \hyperref[p:case1]{1}), factual hallucinations (Cases \hyperref[p:case2]{2}, \hyperref[p:case3]{3}), semantic disarray (Case \hyperref[p:case4]{4}), and
numerical or logical errors (Cases \hyperref[p:case3]{3}, \hyperref[p:case5]{5}), whereas our model accurately predicts outputs, maintaining content faithfulness and structural coherence.

\section{Future Opportunities}
Future work may enhance the text-to-structure generation of agentic AI from multiple directions. First, optimizing real-time conversion with efficient algorithms and lightweight models can support dynamic environments, which is critical for tasks such as question answering and summarization. Second, improving accuracy and consistency through multimodal learning (e.g., integrating text with images) and self-reflection mechanisms can reduce errors in complex texts, thereby benefiting the information state conversion of agentic information retrieval. Third, reinforcement learning with sophisticated reward functions -- balancing utility, efficiency, and ethics -- can combine structured outputs with external goals to enhance autonomous decision-making capabilities. Finally, multi-task learning frameworks can simultaneously generate different structures, such as tables and charts, with consistency and flexibility. These efforts aim to improve the autonomy, accuracy, and applicability of agentic AI in complex real-world scenarios.

\section{Conclusion}
In conclusion, this work aims to provide a comprehensive review of currently existing methods, benchmark datasets, and evaluation metrics for text-to-structure extraction by designing a universal evaluation framework specifically tailored for structured outputs. Through these contributions, we not only synthesize current research progress but also elucidate how these technologies can empower agentic AI systems to achieve enhanced autonomy and efficiency in dynamic real-world environments.

\section*{Limitations}
The limitations of this study mainly stem from two aspects. First, since we strategically focus the most representative studies in this field, the scope of the literature review may not comprehensively cover all research advances related to complex structures. Second, given the rapid development of LLMs, the Text-to-Structure benchmark evaluation may not cover the latest published LLMs. However, despite our best efforts to survey a comprehensive selection of important papers and mainstream LLMs, the practical constraints of a single submission inherently limit our ability to cover every aspect or all LLM variants. Future research directions should aim for a broader literature review while integrating datasets from cross-disciplinary studies to enhance their generalizability.

\section*{Ethics Statement}
Our paper presents a comprehensive survey of text-to-complex structure extraction, with a specific focus on tables, graphs, and charts. The datasets and models employed in this study are all open-source.
All datasets are used in their pre-anonymized forms as released by the original creators, complying with established licensing agreements.
Our evaluation uses strictly unaltered benchmark data, applying only random sampling to select representative subsets. Crucially, no additions, modifications, or external data curation occur at any processing stage, preserving original dataset integrity while enabling efficient LLM assessment. The annotation was strictly limited to validating sampled structural outputs, conducted voluntarily by three doctoral researchers from the authors' institution. Therefore, to the best of the authors' knowledge, we believe that this study introduces no additional risk.

\bibliography{custom}

\appendix
\clearpage
\newpage
\begin{center}
    {\Large\textbf{Appendices}}
\end{center}

\section{Task Definition for Graph Generation} \label{app:xxxx}
This provides an extended definition of graph generation, where graphs can be further categorized into knowledge graphs and mind maps.

\paragraph{Knowledge Graph}
A knowledge graph is a directed graph $\boldsymbol{G}_\text{kg}=(\mathcal{V,E,A,L})$, where:
$\mathcal{V}$ represents entities, $\mathcal{E}$ represents direct relationships, $\mathcal{A}(v)$ and $\mathcal{A}(e)$ are the attributes of entity types and relation types, $\mathcal{L}(v)$ and $\mathcal{L}(e)$ provide contextual meaning for nodes and edges.

\paragraph{Mind Map}
A mind map is a hierarchical directed graph $\boldsymbol{G}_\text{mm}=(\mathcal{V},\mathcal{E},\mathcal{A},\mathcal{L})$, where: $\mathcal{V}$ represents concepts with a unique central node, $\mathcal{E}$ represents directed governing relations, $\mathcal{A}(v)$ distinguish root and subordinate nodes, $\mathcal{L}(v)$ and $\mathcal{L}(e)$ provide contextual meaning for concepts and relationships. There is a hierarchy constraint that each non-root node has a unique path to the root~\cite{DBLP:conf/ijcai/WeiGWS19,DBLP:conf/emnlp/HuGZGS21}.

\section{Detail of Current Datasets} \label{sec:app_datasets}

\subsection{Dataset Categorization} \label{sec:app_dataset_cat}

 We use T2S to differentiate native text-to-structure datasets from those repurposed from structure-to-text tasks, as this fundamental design distinction impacts model performance evaluation. Regarding annotation practices, we consider any human verification or correction, even in crawled datasets, as constituting annotation. Schema constraints distinguish between limited schemas (domain-specific, fixed attributes like sports terminology) and unlimited schemas (open-domain, free-form relations, such as Wikipedia tables). Text inputs are classified by length into sentence, paragraph, and document scales. 
Reasoning complexity indicates the cognitive effort required to transform textual information into structured formats: (1) Low difficulty involves direct extraction of explicit entities/relations, populating a table with stated attributes, or creating a simple bar chart from numeric mentions; (2) Medium difficulty involves context-bound operation, such as resolving coreferences, linking ``it'' to a named entity in a knowledge graph, or aggregating values for chart axes; (3) High difficulty involves global reasoning, for example, temporal reasoning (e.g., constructing timelines from events), or inferring implicit hierarchies (e.g., deducing parent-child relationships in taxonomies for graph structures). The structural complexity can be systematically defined across multiple axes: (1) Tables: low complexity means less than two rows or columns; medium complexity means larger grids  with uniform cell structures; high complexity means complex layouts (e.g., merged cells, nested tables, hierarchical headers); (2) Graphs: the complexity are graded by average node degree and total node count in ascending order; (3) Charts: the complexity are categorized by the diversity of chart types (e.g., bar, line, scatter) in the dataset.
The Gold Test Set (GTS) serves as a critical quality indicator, distinguishing between fully human-annotated test sets and those with alternative provenance. 

\subsection{Common Practices for Data Construction} \label{sec:datasets:built}

\paragraph{LLM Paraphrasing}
During the data generation process, leveraging LLMs for paraphrasing ensures alignment with instructional tone, enhances sentence diversity, and improves the overall data quality~\cite{DBLP:conf/emnlp/Jiao0LZOJ023,DBLP:conf/emnlp/DengC00FZYS24}.

\paragraph{Quality Control}
Various methods have been employed to ensure the quality of custom-built datasets. Some studies randomly select instances for annotation to analyze results~\cite{DBLP:journals/corr/abs-2305-11527,DBLP:conf/naacl/TangZPZZCG24,DBLP:conf/emnlp/DengC00FZYS24}. Others design iterative prompts to ensure data quality~\cite{DBLP:conf/emnlp/LiR024}. Moreover, a combined approach leveraging LLMs and human collaboration has been applied for data cleaning~\cite{DBLP:journals/corr/abs-2405-12174}.

\paragraph{Categorization by Difficulty Level}
It is a common practice that the proposed datasets are pre-divided based on subtask difficulty levels~\cite{DBLP:conf/emnlp/Jiao0LZOJ023,DBLP:conf/emnlp/DengC00FZYS24}. It allows for a more granular assessment of the model's strengths and weaknesses across different complexity levels.

\section{Related Works for Large Language Models}\label{sec:Related_Works_for_LLMs}
Recent comprehensive evaluations have demonstrated that large language models (LLMs) achieve superior performance across numerous natural language processing tasks in zero-shot scenarios~\cite{
DBLP:journals/corr/abs-2303-12712,
DBLP:conf/eacl/ChanCWJFLS24,
DBLP:conf/emnlp/ChengQCFWCRGZSZ23,
DBLP:conf/acl/0001FLS0XWBLJCS24,
DBLP:conf/emnlp/JiayangCZQZLS0L24,
DBLP:conf/ijcnlp/ChanLCCSWS23,
DBLP:conf/emnlp/JiangCCW23,
shi2025inferencedynamicsefficientroutingllms,
DBLP:conf/coling/JiayangQC0SZ24,DBLP:journals/corr/abs-2505-13259}. However, significant challenges persist in complex reasoning domains, including mathematical problem-solving~\cite{DBLP:journals/corr/abs-2301-13867}, theory of mind reasoning~\cite{DBLP:conf/emnlp/ChanJYDF0L0WS24, DBLP:conf/pricai/LinCSL24, DBLP:journals/corr/abs-2506-02461}, analogical reasoning~\cite{DBLP:conf/emnlp/ChengQCFWCRGZSZ23,DBLP:journals/corr/abs-2502-11176,DBLP:journals/corr/abs-2504-05081}, discourse relation classification~\cite{DBLP:conf/acl/ChanLCLSWS23}, shopping intention~\cite{yang2025sessionintentbenchmultitaskintersessionintentionshift}, argument impact classification~\cite{DBLP:conf/ecai/ChanCLYJD0SWS24}, and complex game scenarios~\cite{DBLP:conf/webi/YimCSDFZS24}, alongside important considerations regarding ethical implications and privacy preservation~\cite{DBLP:journals/corr/abs-2212-09292,DBLP:journals/corr/abs-2310-10383,DBLP:journals/corr/abs-2302-00539,DBLP:conf/acl/0003GLFH0CYYS24,DBLP:conf/emnlp/Fan0D0S24, DBLP:conf/aaai/LiCZHCLS25,DBLP:journals/corr/abs-2505-14104}. Most critically for advancing agentic intelligence systems, the capacity of LLMs to transform unstructured text into structured formats—such as tables, knowledge graphs, and charts—remains underexplored despite its fundamental importance for context-aware retrieval and autonomous decision-making processes. This systematic review addresses this critical gap by comprehensively evaluating how LLMs perform across diverse text-to-structure tasks. Through our proposed universal evaluation framework and systematic analysis of current methodologies, we provide the foundation for more reliable and capable agentic intelligence applications.

\section{Details of T2S Benchmark} \label{sec:app_detail_advie}

\begin{figure}[!t]
\begin{tcolorbox}[title={Prompting Template}, colback = cBlue_1!10, colframe = cBlue_6!80,  coltitle=white,fonttitle=\bfseries\small,fontupper=\scriptsize,fontlower=\scriptsize]


\texttt{You are an expert in \{Structure\} evaluation. Your task is to \{Task Description\}. Assign four scores from 0 to 100 based on the following criteria. Please keep the scoring thought process as CONCISE as possible, without repeatedly adjusting the scores.}
\\

\texttt{\# Evaluation Criteria}

$\\$

\texttt{\{Evaluation Criteria\}}

$\\$
\texttt{Please compare the following \{Structure\} and provide me with the final four section scores. Output in this format:}
$\\$

\ \ \ \ \ \ \ \ \texttt{1. **Brief Justification**}  

$\\$

\ \ \ \ \ \ \ \ \texttt{2. **Final Score**}  

$\\$

\ \ \ \ \ \ \ \ \ \ \ \ \ \ \ \ \texttt{Section 1: XX/100}

    $\\$

\ \ \ \ \ \ \ \ \ \ \ \ \ \ \ \ \texttt{Section 2: XX/100}

   $ \\$

\ \ \ \ \ \ \ \ \ \ \ \ \ \ \ \ \texttt{Section 3: XX/100}

   $ \\$

\ \ \ \ \ \ \ \ \ \ \ \ \ \ \ \ \texttt{Section 4: XX/100}

$\\$
\texttt{\# **Ground Truth \{Structure\}**:}

$\\$

\texttt{\{Ground Truth\}}

$\\$
\texttt{\# **LLM-Generated \{Structure\}**:}

$\\$

\texttt{\{LLM Generated Output\}}

$\\$
\texttt{\# **Source Text **:}

$\\$

\texttt{\{Source Text\}}
\end{tcolorbox}
\vspace{-7pt}
    \caption{Prompting template for structure outputs evaluation.}
    \label{fig:template}
\vspace{-10pt}
\end{figure}
\begin{figure*}[h]

\vspace{-0.05in}
\begin{tcolorbox}[title={Evaluation Criteria for Tables}, colback = cBlue_1!10, colframe = cBlue_6!80,  coltitle=white,fonttitle=\bfseries\small,fontupper=\scriptsize,fontlower=\scriptsize]

\begin{enumerate}[topsep=0.2em, parsep=0.1em]
    \item \textbf{Precision (100 points)}
    \begin{itemize}
        \item Evaluates whether generated cells (with its corresponding row/column headers) conflict with ground truth. If no corresponding information exists in the ground truth, compare with the original text. Incorrect information is deducted proportionally from the total score.
        \begin{itemize}
            \item Error Rate = (Number of conflicting cells) / (Total cells)
            \item Final Score = 100 * (1 - Error Rate)
        \end{itemize}
        \item - \textbf{Correctness Guidelines}
      \begin{itemize}
          \item Numerical values: Strict equality required.
          \item Textual content: Semantic equivalence required.
          \item Empty cells: Equivalent to ``not specified'', ``N/A'', or ``-''.
      \end{itemize} 
    \end{itemize}
    \item \textbf{Recall (100 points)}
    \begin{itemize}
        \item For each cell in the ground truth (with its corresponding row/column headers), evaluate the percentage of information captured in the LLM-generated output. Missing or incorrect information is deducted proportionally from the total score.
        \begin{itemize}
            \item Rate = (Number of missing or incorrect cells) / (Total cells)
            \item Final Score = 100 * (1 - Rate)
        \end{itemize}
        \item For \textbf{Correctness Guidelines}, please follow the same criteria as Precision.
    \end{itemize}
    \item \textbf{Alignment (100 points)}
    \begin{itemize}
        \item Evaluates alignment with expected formats, schema relations, and data types (transposed rows/columns allowed without penalty):
        \begin{itemize}
            \item Perfect Match (90-100): Rows/columns fully match reference structure.
            \item Minor Deviations (80-89): Few row/column discrepancies; requires minimal adjustments.
            \item Noticeable Differences (60-79): Multiple added/removed rows/columns; needs manual reorganization.
            \item Severe Deviation (0-59): Missing headers/data type errors; structure unusable.
        \end{itemize}
    \end{itemize}
    
    \item \textbf{Clarity (100 points)}
    \begin{itemize}
        \item Assesses overall intelligibility of the table and clarity and redundancy in headers/cell values:
        \begin{itemize}
            \item Perfect Clarity (90-100): No ambiguity or redundancy.
            \item Minor Issues (80-89): Few redundant/ambiguous terms; core meaning intact.
            \item Significant Issues (60-79): Multiple ambiguous terms requiring inference; extractable core data.
            \item Critical Flaws (0-59): Conflicting headers/unreadable data; table unreliable.
        \end{itemize}
    \end{itemize}
\end{enumerate}

\end{tcolorbox}
\vspace{-0.05in}

\caption{Evaluation Criteria for Tables}
\label{fig:eval_criteria_tables}
\end{figure*}

\begin{figure*}[!h]

\vspace{-0.05in}
\begin{tcolorbox}[title={Evaluation Criteria for Knowledge Graphs}, colback = cBlue_1!10, colframe = cBlue_6!80,  coltitle=white,fonttitle=\bfseries\small,fontupper=\scriptsize,fontlower=\scriptsize]

\begin{enumerate}[topsep=0.2em, parsep=0.1em]
    \item \textbf{Precision (100 points)}
    \begin{itemize}
        \item Evaluates whether generated triples conflict with ground truth. If no corresponding information exists in the ground truth, compare with the original text. Incorrect information is deducted proportionally from the total score.
        \begin{itemize}
            \item Error Rate = (Number of conflicting triples) / (Total triples)
            \item Final Score = 100 * (1 - Error Rate)
        \end{itemize}
        \item - \textbf{Correctness Guidelines}
      \begin{itemize}
          \item Numerical values: Strict equality required.
          \item Textual content: Semantic equivalence required.
      \end{itemize} 
    \end{itemize}
    \item \textbf{Recall (100 points)}
    \begin{itemize}
        \item For each triple in the ground truth, evaluate the percentage of information captured in the LLM-generated triples. Missing or incorrect information is deducted proportionally from the total score.
        \begin{itemize}
            \item Rate = (Number of missing or incorrect triples) / (Total triples)
            \item Final Score = 100 * (1 - Rate)
        \end{itemize}
        \item For \textbf{Correctness Guidelines}, please follow the same criteria as Precision.
    \end{itemize}
    \item \textbf{Alignment (100 points)}
    \begin{itemize}
        \item Evaluates alignment with expected formats, schema relations, and data types (transposed rows/columns allowed without penalty):
        \begin{itemize}
            \item Perfect Match (90-100): Exact schema/format match.
            \item Minor Deviations (80-89):  extra/missing triples; parsable.
            \item Noticeable Differences (60-79): Multiple missing triples/schema gaps; needs manual fixes.
            \item Severe Deviation (0-59): Invalid predicates/scrambled data; KG unusable.
        \end{itemize}
    \end{itemize}
    
    \item \textbf{Clarity (100 points)}
    \begin{itemize}
        \item Assesses clarity, standardization, and absence of ambiguity/redundancy in entities and relations:
        \begin{itemize}
            \item Perfect Clarity (90-100): Standardized terms; zero redundancy.
            \item Minor Issues (80-89): Rare ambiguous labels; core clear.
            \item Significant Issues (60-79): Frequent ambiguous terms; extractable core.
            \item Critical Flaws (0-59): Uninterpretable relations; KG unreliable
        \end{itemize}
    \end{itemize}
\end{enumerate}

\end{tcolorbox}
\vspace{-0.05in}

\caption{Evaluation Criteria for Knowledge Graphs}
\label{fig:eval_criteria_kgs}
\end{figure*}

\begin{figure*}[h]

\vspace{-0.05in}
\begin{tcolorbox}[title={Evaluation Criteria for Charts}, colback = cBlue_1!10, colframe = cBlue_6!80,  coltitle=white,fonttitle=\bfseries\small,fontupper=\scriptsize,fontlower=\scriptsize]

\begin{enumerate}[topsep=0.2em, parsep=0.1em]
    \item \textbf{Precision (100 points)}
    \begin{itemize}
        \item Evaluates whether generated chart elements (data points, labels, categories) conflict with the ground truth. If no ground truth chart exists, compare with the original text.
        \begin{itemize}
            \item Error Rate = (Number of conflicting data points/labels/categories) / (Total data points/labels/categories)
            \item Final Score = 100 * (1 - Error Rate)
        \end{itemize}
        \item - \textbf{Correctness Guidelines}
      \begin{itemize}
          \item Numerical values (e.g., axis values, percentages): Strict equality required (up to two decimal places allowed).
          \item Categorical labels (e.g., axis labels, legend entries): Semantic equivalence required (e.g., "Q1 2023" vs. "First Quarter" is acceptable; "Male" vs. "Female" is not).
          \item  Missing elements: Treat as errors if present in the ground truth but omitted in the generated chart.
      \end{itemize} 
    \end{itemize}
    \item \textbf{Recall (100 points)}
    \begin{itemize}
        \item Measures how well the generated chart captures all data points, trends, and categories from the ground truth.
        \begin{itemize}
            \item Rate = (Number of missing or misrepresented data points/categories/labels) / (Total data points/categories/labels)
            \item Final Score = 100 * (1 - Rate)
        \end{itemize}
        \item For \textbf{Correctness Guidelines}, please follow the same criteria as Precision, prioritizing critical elements (e.g., peaks, outliers, primary categories).
    \end{itemize}
    \item \textbf{Alignment (100 points)}
    \begin{itemize}
        \item Alignment with expected chart structure:
        \begin{itemize}
            \item Perfect (90-100): Chart type, headers/labels, and groupings fully match reference.
            \item Minor flaws (80-89): Formatting/positioning deviations (e.g., axis units, legend placement) with minimal impact.
            \item Major errors (60-79): Incorrect chart type, missing key labels, or misaligned data groups requiring manual fixes.
            \item Invalid (0-59): Broken structure (e.g., inverted axes, data-trend conflicts).
        \end{itemize}
    \end{itemize}
    
    \item \textbf{Clarity (100 points)}
    \begin{itemize}
        \item Clarity and truthfulness:
        \begin{itemize}
            \item Perfect (90-100): Visual encodings (color/size/position) logically match data; no ambiguity.
            \item Minor issues (80-89): Non-critical redundancy/design flaws (e.g., label overlaps) with preserved meaning.
            \item Confusing (60-79): Ambiguous scales/axes requiring user guesses.
            \item Misleading (0-59): Distorted representations (e.g., truncated axes, false color mappings) altering insights.
        \end{itemize}
    \end{itemize}
\end{enumerate}

\end{tcolorbox}
\vspace{-0.05in}

\caption{Evaluation Criteria for Charts}
\label{fig:eval_criteria_charts}
\end{figure*}

\subsection{Details of Evaluation Criteria} \label{sec:app_criteria_detail}
As mentioned in Section~\ref{sec:score_criteria}, we propose an evaluation framework consisting of content faithfulness (precision and recall) and structure coherence (alignment and clarity). To integrate these metrics into the context of agentic AI systems, we assess their application across four key aspects of text-to-structure generation, which is a critical process for enhancing agents' autonomy and effectiveness. Figure~\ref{fig:template} shows the prompting template for the evaluation of structure outputs. For the complete version of the evaluation criteria regarding tables, graphs, and charts in the prompting template, please refer to Figure~\ref{fig:eval_criteria_tables}, \ref{fig:eval_criteria_kgs}, \ref{fig:eval_criteria_charts}, respectively. Below, we outline how each metric applies to this process:

\paragraph{Precision} 
Precision evaluates the structured outputs by identifying conflicts with the ground truth and source text, such as factual errors or mismatches. In applications like automated report generation or document extraction, precision ensures that the AI-generated tables and charts do not introduce mistakes. The detailed calculation rules for precision are outlined in the prompt -- it is defined as the percentage of incorrect information relative to the total generated information, multiplied by 100 to obtain the precision score. When comparing whether two pieces of information are equivalent, numerical values, textual content, and other data types are evaluated separately.
\paragraph{Recall}
Recall evaluates the completeness of the structured outputs, assessing how thoroughly the AI captures details from the source text. It is crucial for ensuring that no critical information is omitted during the text-to-structure conversion. The detailed calculation rules for recall are detailed in the prompt -- it is determined by the percentage of information from the ground truth that appears in the output, multiplied by 100 to obtain the recall score.

\paragraph{Alignment}
Alignment evaluates how well the structured outputs conform to expected formats, such as consistent schemas or data types, ensuring seamless processing by the AI or downstream applications. The detailed calculation rules for alignment are outlined in the prompt -- it is generally assessed by categorizing the degree of alignment between the output and the instruction across expected formats, data types, and other relevant dimensions into four scoring tiers.

\paragraph{Clarity}
Clarity evaluates the logical organization and presentation of the structured outputs, ensuring they are unambiguous and free of redundancy. This will enhance the efficiency of data interpretation and avoid confusion to maintain usability. The detailed calculation rules for clarity are outlined in the prompt -- the outputs will be evaluated and scored across four tiers based on degree of ambiguity, comprehensibility, and redundancy. 

\noindent 
Taking applying to agentic workflows as an example, this scoring framework optimizes the performance of text-to-structure generation: \textit{precision} and \textit{recall} validate extracted data fidelity, ensuring agents develop an accurate and comprehensive understanding of data, thereby facilitating the generation of reliable and holistic decisions; while \textit{alignment} and \textit{clarity} govern integration with downstream modules, enhancing interoperability within agentic AI systems and optimizing the efficiency of data interpretation and action.

\subsection{Selection of Datasets}\label{sec:app_dataset_selection}
Regarding the selection of datasets for our Text-to-Structure benchmark, we first ensure that the chosen datasets are open-source and that their test sets are manually annotated to guarantee accuracy. For the text-to-table generation task, we select three datasets, each with distinct characteristics:
\begin{itemize}
    \item \textbf{InstructIE}~\cite{DBLP:conf/emnlp/Jiao0LZOJ023}: Each (text, table) pair is accompanied by a complex instruction, with a unique instruction for different instances.
    \item 
    \textbf{Struc-Bench Table}~\cite{DBLP:conf/naacl/TangZPZZCG24}: A cleaned version of RotoWire dataset~\cite{DBLP:conf/emnlp/WisemanSR17}, serving as a traditional benchmark since the beginning of text-to-table tasks.
    \item \textbf{LiveSum}~\cite{DBLP:conf/emnlp/DengC00FZYS24}: Focuses on evaluating a model's ability to integrate information from text into tables, a capability currently lacking in many LLMs.
\end{itemize}
For the text-to-graph generation task, the available datasets are limited. We select two open-domain datasets: \textbf{DART}~\cite{DBLP:conf/naacl/NanRZRSHTVVKLIP21} and \textbf{InstructIE}~\cite{DBLP:journals/corr/abs-2305-11527}, where the latter includes half of its data in Chinese, enabling evaluation of multilingual performance. In the text-to-chart generation task, due to the scarcity of dedicated datasets, we choose \textbf{Text2Chart31}~\cite{DBLP:conf/emnlp/ZadehKKK24}, the most comprehensive dataset in terms of chart type coverage, which effectively tests the generalization capability of models. To standardize the datasets, we fix the training set sample size to 1,000 instances during SFT and sample 250 instances from each dataset's test set for evaluation.

\subsection{Selection of Models} \label{sec:app_select_models}
The selected LLMs in our study can be categorized into two types: non-thinking models and thinking models. Specifically, the non-thinking models include: Mistral-7B~\cite{DBLP:journals/corr/abs-2310-06825}, Llama-3.1-8B~\cite{DBLP:journals/corr/abs-2407-21783}, R1-Distill-Llama-8B~\cite{deepseekai2025deepseekr1incentivizingreasoningcapability}, Phi-3-medium~\cite{abdin2024phi3technicalreporthighly}, Mixtral-8x22B~\cite{jiang2024mixtralexperts}, Llama-3.1-405B~\cite{DBLP:journals/corr/abs-2407-21783}, Llama-3.3-70B~\cite{DBLP:journals/corr/abs-2407-21783}, GPT-3.5-turbo~\cite{openaichatgpt} and GPT-4o~\cite{openai2024gpt4}; while the thinking models comprise QwQ-32B~\cite{qwq32b}, DeepSeek-R1~\cite{deepseekai2025deepseekr1incentivizingreasoningcapability}, Grok-3-mini~\cite{grok3mini}, and O4-mini~\cite{openaio4mini}. For SFT experiments, we evaluate models Mistral-7B, Llama-3.1-8B, R1-Distill-Llama-8B. The remaining models are assessed under a zero-shot setting. Additionally, for non-thinking models, we further examine their performance when augmented with CoT prompting. For the SFT models, our experiments are conducted on a single H20 GPU card. For other LLMs, we access their capabilities through API calls. Specifically, we conduct model inference with temperature set to 0 for non-thinking models, while adopting the officially recommended hyperparameters for thinking models as specified in their respective documentation, with all results averaged over five independent runs. When employing DeepSeek-R1 and GPT-4o as LLM-as-judge, we fix the temperature parameter at 0 to ensure that every evaluation remains deterministic.

\begin{table*}[!t]
    \centering
    \renewcommand\arraystretch{1.0}
    \scalebox{0.66}{
    \begin{tabular}{p{3.6cm}p{0.7cm}p{0.7cm}p{0.7cm}p{0.7cm}p{0.7cm}p{0.7cm}p{0.7cm}p{0.7cm}p{0.7cm}p{0.7cm}p{0.7cm}p{0.7cm}p{0.7cm}p{0.7cm}p{0.7cm}p{0.7cm}p{0.7cm}p{0.7cm}}
        \toprule
        \multirow{3}{*}{\textbf{Model}}  
        & \multicolumn{6}{c}{\textbf{InstructIE}} &  \multicolumn{6}{c}{\textbf{Struc-Bench}} & \multicolumn{6}{c}{\textbf{LiveSum}} \\
        & \multicolumn{6}{c}{{\footnotesize \cite{DBLP:conf/emnlp/Jiao0LZOJ023}}} &  \multicolumn{6}{c}{\footnotesize \cite{DBLP:conf/naacl/TangZPZZCG24}} & \multicolumn{6}{c}{\footnotesize \cite{DBLP:conf/emnlp/DengC00FZYS24}} \\
         \cmidrule(r){2-7} \cmidrule(r){8-13} \cmidrule(r){14-19} 
         & \footnotesize Prec.  & \footnotesize Rec. & \footnotesize \textsc{\textbf{Faith}}. & \footnotesize Align.  & \footnotesize Clr. &\footnotesize \textsc{\textbf{Coh}.}& \footnotesize Prec.  & \footnotesize Rec. & \footnotesize \textsc{\textbf{Faith}}. & \footnotesize Align.  & \footnotesize Clr. &\footnotesize \textsc{\textbf{Coh}.}& \footnotesize Prec.  & \footnotesize Rec. & \footnotesize \textsc{\textbf{Faith}}. & \footnotesize Align.  & \footnotesize Clr. &\footnotesize \textsc{\textbf{Coh}.}       \\
        \midrule
        \multicolumn{19}{l}{$\vardiamondsuit$ \emph{\textbf{SFT}}} \\
Mistral-7B  & \textcolor{above60}{67.84} & \textcolor{below60}{55.93} & \textbf{\textcolor{below60}{58.08}} &  \textcolor{above60}{67.02} & \textcolor{above70}{74.60} & \textbf{\textcolor{above60}{68.85}}  & \underline{\textcolor{above90}{91.14}} & \textcolor{above80}{86.73} & \textbf{\textcolor{above80}{87.65}} &  \textcolor{above90}{95.47} & \textcolor{above80}{87.62} & \textbf{\textcolor{above90}{90.05}}  & \textcolor{below60}{53.11} & \textcolor{below60}{53.87} & \textbf{\textcolor{below60}{53.31}} &  \underline{\textcolor{above90}{99.81}} & \textcolor{above90}{97.78} & \textbf{\textcolor{above90}{98.54}} \\
Llama-3.1-8B  & \textcolor{above60}{64.38} & \textcolor{above60}{60.34} & \textbf{\textcolor{below60}{59.31}} &  \textcolor{above60}{64.61} & \textcolor{above70}{72.28} & \textbf{\textcolor{above60}{66.44}}  & \textcolor{above80}{87.02} & \textcolor{above80}{83.94} & \textbf{\textcolor{above80}{85.01}} &  \textcolor{above90}{90.38} & \textcolor{above80}{82.67} & \textbf{\textcolor{above80}{85.95}}  & \textcolor{above60}{61.69} & \textcolor{above60}{61.46} & \textbf{\textcolor{above60}{61.34}} &  \textcolor{above90}{90.56} & \textcolor{above90}{90.65} & \textbf{\textcolor{above90}{90.30}} \\
R1-Distill-Llama-8B  & \textcolor{above70}{70.39} & \textcolor{above60}{61.44} & \textbf{\textcolor{above60}{62.86}} &  \textcolor{above60}{64.69} & \textcolor{above70}{75.12} & \textbf{\textcolor{above60}{68.08}}  & \textcolor{above60}{63.97} & \textcolor{below60}{54.84} & \textbf{\textcolor{below60}{54.56}} &  \textcolor{above60}{62.55} & \textcolor{below60}{57.34} & \textbf{\textcolor{below60}{58.97}}  & \textcolor{below60}{52.93} & \textcolor{below60}{43.21} & \textbf{\textcolor{below60}{43.27}} &  \textcolor{above80}{86.28} & \textcolor{above80}{86.44} & \textbf{\textcolor{above80}{86.22}} \\
        \midrule
        \multicolumn{19}{l}{$\vardiamondsuit$ \emph{\textbf{Zero-Shot}}} \\
Phi-3-medium  & \textcolor{below60}{43.19} & \textcolor{below60}{28.79} & \textbf{\textcolor{below60}{29.15}} &  \textcolor{below60}{40.42} & \textcolor{below60}{42.75} & \textbf{\textcolor{below60}{39.89}}  & \textcolor{above70}{74.68} & \textcolor{above70}{75.36} & \textbf{\textcolor{above70}{74.04}} &  \textcolor{above90}{91.79} & \textcolor{above70}{78.68} & \textbf{\textcolor{above80}{83.99}}  & \textcolor{below60}{41.84} & \textcolor{below60}{42.27} & \textbf{\textcolor{below60}{41.61}} &  \textcolor{above90}{93.64} & \textcolor{above90}{90.62} & \textbf{\textcolor{above90}{91.73}} \\
Phi-3-medium~(\textsc{CoT})  & \textcolor{below60}{42.85} & \textcolor{below60}{29.82} & \textbf{\textcolor{below60}{30.36}} &  \textcolor{below60}{37.83} & \textcolor{below60}{39.56} & \textbf{\textcolor{below60}{37.23}}  & \textcolor{above70}{74.91} & \textcolor{above70}{73.83} & \textbf{\textcolor{above70}{73.42}} &  \textcolor{above80}{88.81} & \textcolor{above70}{75.92} & \textbf{\textcolor{above80}{81.06}}  & \textcolor{below60}{46.57} & \textcolor{below60}{46.59} & \textbf{\textcolor{below60}{46.36}} &  \textcolor{above90}{94.40} & \textcolor{above90}{92.54} & \textbf{\textcolor{above90}{93.19}} \\
Mistral-7B  & \textcolor{below60}{58.53} & \textcolor{below60}{53.29} & \textbf{\textcolor{below60}{53.00}} &  \textcolor{below60}{57.05} & \textcolor{above60}{65.53} & \textbf{\textcolor{below60}{59.73}}  & \textcolor{below60}{55.17} & \textcolor{above60}{61.17} & \textbf{\textcolor{below60}{56.19}} &  \textcolor{above70}{72.52} & \textcolor{above60}{64.05} & \textbf{\textcolor{above60}{67.05}}  & \textcolor{below60}{31.25} & \textcolor{below60}{30.56} & \textbf{\textcolor{below60}{30.14}} &  \textcolor{above90}{90.48} & \textcolor{above80}{85.78} & \textbf{\textcolor{above80}{87.26}} \\
Mistral-7B~(\textsc{CoT})  & \textcolor{below60}{59.60} & \textcolor{below60}{53.82} & \textbf{\textcolor{below60}{54.02}} &  \textcolor{below60}{57.30} & \textcolor{above60}{66.83} & \textbf{\textcolor{above60}{60.37}}  & \textcolor{below60}{54.35} & \textcolor{above60}{62.38} & \textbf{\textcolor{below60}{56.25}} &  \textcolor{above70}{76.51} & \textcolor{above60}{66.44} & \textbf{\textcolor{above70}{70.34}}  & \textcolor{below60}{42.38} & \textcolor{below60}{42.14} & \textbf{\textcolor{below60}{41.27}} &  \textcolor{above80}{89.75} & \textcolor{above80}{83.43} & \textbf{\textcolor{above80}{85.97}} \\
Mixtral-8x22B  & \textcolor{above60}{66.92} & \textcolor{below60}{54.34} & \textbf{\textcolor{below60}{56.35}} &  \textcolor{above60}{66.39} & \textcolor{above70}{73.42} & \textbf{\textcolor{above60}{68.25}}  & \textcolor{above80}{84.95} & \textcolor{above80}{84.71} & \textbf{\textcolor{above80}{84.16}} &  \textcolor{above80}{89.00} & \textcolor{above80}{82.74} & \textbf{\textcolor{above80}{85.20}}  & \textcolor{below60}{49.74} & \textcolor{below60}{48.97} & \textbf{\textcolor{below60}{49.12}} &  \textcolor{above80}{87.54} & \textcolor{above80}{88.91} & \textbf{\textcolor{above80}{87.91}} \\
Mixtral-8x22B~(\textsc{CoT})  & \textcolor{above60}{66.82} & \textcolor{below60}{56.05} & \textbf{\textcolor{below60}{58.27}} &  \textcolor{above60}{64.97} & \textcolor{above70}{74.73} & \textbf{\textcolor{above60}{68.29}}  & \textcolor{above80}{82.81} & \textcolor{above80}{80.26} & \textbf{\textcolor{above80}{80.55}} &  \textcolor{above70}{74.98} & \textcolor{above70}{75.91} & \textbf{\textcolor{above70}{74.94}}  & \textcolor{below60}{55.68} & \textcolor{below60}{54.43} & \textbf{\textcolor{below60}{54.87}} &  \textcolor{above80}{86.75} & \textcolor{above80}{88.51} & \textbf{\textcolor{above80}{87.37}} \\
Llama-3.1-405B  & \textcolor{above70}{71.31} & \textcolor{above60}{62.27} & \textbf{\textcolor{above60}{63.88}} &  \textcolor{above60}{65.36} & \textcolor{above70}{77.31} & \textbf{\textcolor{above60}{69.63}}  & \textcolor{above90}{90.20} & \underline{\textcolor{above80}{89.32}} & \textbf{\underline{\textcolor{above80}{89.43}}} &  \textcolor{above90}{91.57} & \textcolor{above90}{90.49} & \textbf{\textcolor{above90}{90.43}}  & \textcolor{above60}{62.46} & \textcolor{above60}{62.38} & \textbf{\textcolor{above60}{62.29}} &  \textcolor{above90}{99.55} & \textcolor{above90}{98.94} & \textbf{\textcolor{above90}{99.11}} \\
Llama-3.1-405B~(\textsc{CoT})  & \textcolor{above70}{72.43} & \textcolor{above60}{62.39} & \textbf{\textcolor{above60}{64.54}} &  \textcolor{above60}{69.43} & \textcolor{above70}{78.69} & \textbf{\textcolor{above70}{72.80}}  & \textcolor{above90}{90.69} & \textcolor{above80}{88.69} & \textbf{\textcolor{above80}{89.38}} &  \textcolor{above90}{91.72} & \underline{\textcolor{above90}{90.51}} & \textbf{\textcolor{above90}{90.74}}  & \textcolor{above60}{66.27} & \textcolor{above60}{66.33} & \textbf{\textcolor{above60}{66.25}} &  \textcolor{above90}{98.87} & \textcolor{above90}{98.51} & \textbf{\textcolor{above90}{98.64}} \\
Llama-3.3-70B  & \textcolor{above70}{71.10} & \textcolor{above60}{60.09} & \textbf{\textcolor{above60}{62.88}} &  \textcolor{above60}{64.63} & \textcolor{above70}{76.50} & \textbf{\textcolor{above60}{69.19}}  & \textcolor{above80}{82.00} & \textcolor{above80}{80.98} & \textbf{\textcolor{above80}{80.58}} &  \textcolor{above80}{84.68} & \textcolor{above70}{77.58} & \textbf{\textcolor{above70}{80.00}}  & \textcolor{below60}{58.55} & \textcolor{below60}{58.63} & \textbf{\textcolor{below60}{58.48}} &  \textcolor{above90}{99.40} & \underline{\textcolor{above90}{99.13}} & \textbf{\underline{\textcolor{above90}{99.19}}} \\
Llama-3.3-70B~(\textsc{CoT})  & \textcolor{above70}{71.71} & \textcolor{above60}{63.35} & \textbf{\textcolor{above60}{65.34}} &  \textcolor{above60}{64.79} & \textcolor{above70}{77.99} & \textbf{\textcolor{above60}{69.46}}  & \textcolor{above80}{87.53} & \textcolor{above80}{86.47} & \textbf{\textcolor{above80}{86.47}} &  \textcolor{above80}{87.16} & \textcolor{above80}{85.71} & \textbf{\textcolor{above80}{85.84}}  & \textcolor{above60}{62.66} & \textcolor{above60}{63.00} & \textbf{\textcolor{above60}{62.79}} &  \textcolor{above90}{98.98} & \textcolor{above90}{98.47} & \textbf{\textcolor{above90}{98.68}} \\
GPT-3.5-turbo  & \textcolor{above60}{65.29} & \textcolor{below60}{53.80} & \textbf{\textcolor{below60}{55.09}} &  \textcolor{above60}{67.05} & \textcolor{above70}{74.99} & \textbf{\textcolor{above60}{69.01}}  & \textcolor{above70}{79.60} & \textcolor{above70}{79.61} & \textbf{\textcolor{above70}{78.44}} &  \textcolor{above90}{92.09} & \textcolor{above80}{81.98} & \textbf{\textcolor{above80}{85.37}}  & \textcolor{below60}{46.08} & \textcolor{below60}{45.39} & \textbf{\textcolor{below60}{45.38}} &  \textcolor{above70}{77.44} & \textcolor{above80}{82.92} & \textbf{\textcolor{above70}{79.18}} \\
GPT-3.5-turbo~(\textsc{CoT})  & \textcolor{above60}{68.95} & \textcolor{below60}{55.61} & \textbf{\textcolor{below60}{58.44}} &  \textcolor{above60}{65.69} & \textcolor{above70}{75.21} & \textbf{\textcolor{above60}{68.80}}  & \textcolor{above70}{72.86} & \textcolor{above70}{75.03} & \textbf{\textcolor{above70}{71.88}} &  \textcolor{above80}{89.55} & \textcolor{above70}{76.30} & \textbf{\textcolor{above80}{81.54}}  & \textcolor{below60}{45.14} & \textcolor{below60}{45.57} & \textbf{\textcolor{below60}{45.09}} &  \textcolor{above80}{85.88} & \textcolor{above80}{87.71} & \textbf{\textcolor{above80}{86.37}} \\
GPT-4o  & \textcolor{above70}{75.17} & \textcolor{above70}{71.67} & \textbf{\textcolor{above70}{70.72}} &  \underline{\textcolor{above70}{72.97}} & \textcolor{above80}{82.35} & \textbf{\underline{\textcolor{above70}{76.36}}}  & \textcolor{above80}{88.00} & \textcolor{above80}{87.44} & \textbf{\textcolor{above80}{87.35}} &  \underline{\textcolor{above90}{96.34}} & \textcolor{above80}{89.22} & \textbf{{\textcolor{above90}{92.22}}}  & \textcolor{above60}{63.46} & \textcolor{above60}{63.09} & \textbf{\textcolor{above60}{63.21}} &  \textcolor{above90}{91.46} & \textcolor{above90}{92.96} & \textbf{\textcolor{above90}{91.99}} \\
GPT-4o~(\textsc{CoT})  & \textcolor{above70}{75.91} & \textcolor{above60}{67.93} & \textbf{\textcolor{above60}{68.11}} &  \textcolor{above70}{71.97} & \textcolor{above80}{80.15} & \textbf{\textcolor{above70}{74.40}}  & \textcolor{above80}{87.37} & \textcolor{above80}{85.46} & \textbf{\textcolor{above80}{86.05}} &  \textcolor{above90}{94.29} & \textcolor{above80}{87.24} & \textbf{\textcolor{above90}{90.03}}  & \textcolor{above60}{61.96} & \textcolor{above60}{62.88} & \textbf{\textcolor{above60}{62.16}} &  \textcolor{above90}{98.76} & \textcolor{above90}{98.38} & \textbf{\textcolor{above90}{98.50}} \\
QwQ-32B & \textcolor{above70}{75.53} & \textcolor{above60}{66.66} & \textcolor{above60}{\textbf{68.23}} & \textcolor{above60}{63.08} & \textcolor{above70}{77.90} & \textcolor{above60}{\textbf{68.79}} & \textcolor{above80}{81.15} & \textcolor{above80}{81.29} & \textcolor{above80}{\textbf{81.21}} & \textcolor{above90}{90.97} & \textcolor{above80}{84.03} & \textcolor{above80}{\textbf{87.12}} & \textcolor{below60}{58.72} & \textcolor{below60}{58.47} & \textcolor{below60}{\textbf{58.06}} & \textcolor{above90}{95.94} & \textcolor{above90}{96.95} & \textcolor{above90}{\textbf{96.20}} \\
DeepSeek-R1  & \textcolor{above70}{79.05} & \underline{\textcolor{above70}{78.72}} & \underline{\textcolor{above70}{\textbf{76.86}}} & \textcolor{above70}{70.04} & {\textcolor{above80}{82.93}} & \textcolor{above70}{\textbf{74.75}} & \textcolor{above80}{88.19} & \textcolor{above80}{87.95} & \textcolor{above80}{\textbf{87.77}} & \textcolor{above90}{94.79} & \textcolor{above80}{89.32} & \textcolor{above90}{\textbf{91.05}} & \textcolor{above60}{64.37} & \textcolor{above60}{64.79} & \textcolor{above60}{\textbf{64.89}} & \textcolor{above90}{96.74} & \textcolor{above90}{97.88} & \textcolor{above90}{\textbf{96.85}} \\
Grok-3-mini & \underline{\textcolor{above80}{{81.34}}} & \textcolor{above70}{72.42} & \textcolor{above70}{\textbf{75.57}} & \textcolor{above60}{68.93} & \textcolor{above80}{82.91} & \textcolor{above70}{\textbf{75.40}} & \textcolor{above80}{86.17} & \textcolor{above80}{85.45} & \textcolor{above80}{\textbf{85.60}} & \textcolor{above90}{94.61} & \textcolor{above80}{89.59} & \underline{\textcolor{above90}{\textbf{92.28}}} & \textcolor{above60}{65.81} & \textcolor{above60}{65.83} & \textcolor{above60}{\textbf{66.35}} & \textcolor{above90}{96.45} & \textcolor{above90}{94.27} & \textcolor{above90}{\textbf{95.26}} \\
O4-mini & \textcolor{above80}{80.86} & \textcolor{above70}{73.26} & \textcolor{above70}{\textbf{75.71}} & \textcolor{above60}{68.73} & \underline{\textcolor{above80}{83.57}} & \textcolor{above70}{\textbf{74.85}} & \textcolor{above80}{84.13} & \textcolor{above80}{84.83} & \textcolor{above80}{\textbf{83.45}} & \textcolor{above90}{94.52} & \textcolor{above80}{86.30} & \textcolor{above90}{\textbf{90.02}} & \underline{\textcolor{above80}{84.20}} & \underline{\textcolor{above80}{84.43}} & \underline{\textcolor{above80}{\textbf{84.48}}} & \textcolor{above90}{98.64} & \textcolor{above90}{98.65} & \textcolor{above90}{\textbf{99.16}} \\

\bottomrule
    \end{tabular}}
    \caption{Performance of different models on three text-to-table datasets, with the highest scores in each metric \underline{underlined}.} 
    \label{tab:text-to-table}
    \vspace{-10pt}
\end{table*}

\begin{table*}[!t]
    \centering
    \renewcommand\arraystretch{1.0}
    \scalebox{0.75}{
    \begin{tabular}{p{3.6cm}P{1.0cm}P{1.0cm}P{1.0cm}P{1.0cm}P{1.0cm}P{1.0cm}P{1.0cm}P{1.0cm}P{1.0cm}P{1.0cm}P{1.0cm}P{1.0cm}}
        \toprule
        \multirow{3}{*}{\textbf{Model}} & \multicolumn{6}{c}{\textbf{DART}} &  \multicolumn{6}{c}{\textbf{InstructIE}} \\
        & \multicolumn{6}{c}{{\footnotesize \cite{DBLP:conf/naacl/NanRZRSHTVVKLIP21}}} &  \multicolumn{6}{c}{\footnotesize \cite{DBLP:journals/corr/abs-2305-11527}} \\
         \cmidrule(r){2-7} \cmidrule(r){8-13} 
         & \footnotesize Prec.  & \footnotesize Rec. & \footnotesize \textsc{\textbf{Faith}}. & \footnotesize Align.  & \footnotesize Clr. &\footnotesize \textsc{\textbf{Coh}.}& \footnotesize Prec.  & \footnotesize Rec. & \footnotesize \textsc{\textbf{Faith}}. & \footnotesize Align.  & \footnotesize Clr. &\footnotesize \textsc{\textbf{Coh}.}    \\
        \midrule
        \multicolumn{13}{l}{$\vardiamondsuit$ \emph{\textbf{SFT}}} \\
Mistral-7B  & \textcolor{above80}{80.99} & \textcolor{above70}{70.81} & \textbf{\textcolor{above70}{75.07}} &  \underline{\textcolor{above80}{83.15}} & \underline{\textcolor{above80}{83.54}} & \textbf{\underline{\textcolor{above80}{83.17}}}  & \textcolor{below60}{59.75} & \textcolor{below60}{45.35} & \textbf{\textcolor{below60}{44.58}} &  \textcolor{above70}{72.21} & \textcolor{above60}{69.31} & \textbf{\textcolor{above60}{69.67}} \\
Llama-3.1-8B  & \textcolor{above60}{63.63} & \textcolor{below60}{35.92} & \textbf{\textcolor{below60}{38.94}} &  \textcolor{below60}{55.56} & \textcolor{below60}{56.75} & \textbf{\textcolor{below60}{55.13}}  & \textcolor{below60}{55.37} & \textcolor{below60}{35.73} & \textbf{\textcolor{below60}{34.93}} &  \textcolor{above60}{66.42} & \textcolor{above60}{65.70} & \textbf{\textcolor{above60}{64.99}} \\
R1-Distill-Llama-8B  & \textcolor{above60}{68.98} & \textcolor{below60}{54.83} & \textbf{\textcolor{below60}{59.03}} &  \textcolor{above60}{68.41} & \textcolor{above70}{71.40} & \textbf{\textcolor{above60}{69.43}}  & \textcolor{above70}{71.78} & \textcolor{below60}{33.35} & \textbf{\textcolor{below60}{37.27}} &  \textcolor{above70}{70.89} & \textcolor{above60}{69.91} & \textbf{\textcolor{above60}{69.73}} \\
        \midrule
        \multicolumn{13}{l}{$\vardiamondsuit$ \emph{\textbf{Zero-Shot}}} \\
Phi-3-medium  & \textcolor{above70}{73.80} & \textcolor{above60}{60.80} & \textbf{\textcolor{above60}{64.60}} &  \textcolor{above70}{72.36} & \textcolor{above70}{73.26} & \textbf{\textcolor{above70}{72.59}}  & \textcolor{above70}{74.64} & \textcolor{below60}{39.84} & \textbf{\textcolor{below60}{43.40}} &  \textcolor{above70}{74.08} & \textcolor{above70}{71.68} & \textbf{\textcolor{above70}{71.44}} \\
Phi-3-medium~(\textsc{CoT})  & \textcolor{above70}{73.69} & \textcolor{above60}{61.10} & \textbf{\textcolor{above60}{64.88}} &  \textcolor{above70}{71.77} & \textcolor{above70}{73.52} & \textbf{\textcolor{above70}{72.32}}  & \textcolor{above70}{74.54} & \textcolor{below60}{39.10} & \textbf{\textcolor{below60}{41.28}} &  \textcolor{above70}{74.39} & \textcolor{above70}{70.40} & \textbf{\textcolor{above70}{70.44}} \\
Mistral-7B  & \textcolor{above70}{74.22} & \textcolor{above60}{69.81} & \textbf{\textcolor{above70}{70.27}} &  \textcolor{above70}{73.34} & \textcolor{above70}{73.44} & \textbf{\textcolor{above70}{73.06}}  & \textcolor{above60}{65.88} & \textcolor{below60}{41.21} & \textbf{\textcolor{below60}{42.74}} &  \textcolor{above60}{69.34} & \textcolor{above60}{66.85} & \textbf{\textcolor{above60}{66.54}} \\
Mistral-7B~(\textsc{CoT})  & \textcolor{above70}{76.98} & \textcolor{below60}{59.92} & \textbf{\textcolor{above60}{64.68}} &  \textcolor{above70}{70.70} & \textcolor{above70}{71.96} & \textbf{\textcolor{above70}{70.96}}  & \textcolor{above60}{66.78} & \textcolor{below60}{30.86} & \textbf{\textcolor{below60}{33.35}} &  \textcolor{above60}{66.36} & \textcolor{above60}{67.40} & \textbf{\textcolor{above60}{66.01}} \\
Mixtral-8x22B  & \textcolor{above70}{70.58} & \textcolor{above60}{65.30} & \textbf{\textcolor{above60}{65.70}} &  \textcolor{above70}{71.72} & \textcolor{above70}{71.07} & \textbf{\textcolor{above70}{71.07}}  & \textcolor{above60}{68.69} & \textcolor{below60}{38.65} & \textbf{\textcolor{below60}{40.07}} &  \textcolor{above70}{71.32} & \textcolor{above60}{68.57} & \textbf{\textcolor{above60}{68.89}} \\
Mixtral-8x22B~(\textsc{CoT})  & \textcolor{above70}{78.14} & \textcolor{above60}{69.19} & \textbf{\textcolor{above70}{71.27}} &  \textcolor{above70}{74.93} & \textcolor{above70}{74.43} & \textbf{\textcolor{above70}{74.53}}  & \textcolor{above70}{75.68} & \textcolor{below60}{43.30} & \textbf{\textcolor{below60}{47.47}} &  \textcolor{above70}{72.92} & \textcolor{above70}{71.31} & \textbf{\textcolor{above70}{71.29}} \\
Llama-3.1-405B  & \textcolor{above70}{75.77} & \textcolor{above60}{65.24} & \textbf{\textcolor{above60}{68.74}} &  \textcolor{above70}{75.59} & \textcolor{above70}{76.11} & \textbf{\textcolor{above70}{75.61}}  & \textcolor{above70}{73.79} & \textcolor{below60}{43.30} & \textbf{\textcolor{below60}{46.72}} &  \textcolor{above70}{78.16} & \textcolor{above70}{75.52} & \textbf{\textcolor{above70}{75.82}} \\
Llama-3.1-405B~(\textsc{CoT})  & \textcolor{above80}{83.56} & \textcolor{above60}{66.49} & \textbf{\textcolor{above70}{72.31}} &  \textcolor{above70}{77.99} & \textcolor{above70}{78.57} & \textbf{\textcolor{above70}{78.09}}  & \textcolor{above70}{79.32} & \textcolor{below60}{43.46} & \textbf{\textcolor{below60}{47.94}} &  \textcolor{above70}{79.90} & \textcolor{above70}{78.10} & \textbf{\textcolor{above70}{78.48}} \\
Llama-3.3-70B  & \textcolor{above70}{79.22} & \textcolor{above60}{66.95} & \textbf{\textcolor{above70}{71.09}} &  \textcolor{above70}{78.34} & \textcolor{above70}{77.21} & \textbf{\textcolor{above70}{77.54}}  & \textcolor{above70}{77.23} & \textcolor{below60}{44.64} & \textbf{\textcolor{below60}{47.11}} &  \textcolor{above70}{78.70} & \textcolor{above70}{76.36} & \textbf{\textcolor{above70}{76.30}} \\
Llama-3.3-70B~(\textsc{CoT})  & \textcolor{above80}{85.74} & \textcolor{above70}{71.69} & \textbf{\textcolor{above70}{77.11}} &  \textcolor{above70}{79.02} & \textcolor{above70}{79.24} & \textbf{\textcolor{above70}{78.99}}  & \textcolor{above70}{76.47} & \textcolor{below60}{41.80} & \textbf{\textcolor{below60}{46.56}} &  \textcolor{above80}{80.23} & \textcolor{above70}{78.08} & \textbf{\textcolor{above70}{78.51}} \\
GPT-3.5-turbo  & \textcolor{above80}{80.95} & \textcolor{above60}{64.02} & \textbf{\textcolor{above60}{69.75}} &  \textcolor{above70}{76.56} & \textcolor{above70}{77.27} & \textbf{\textcolor{above70}{76.72}}  & \textcolor{above70}{74.63} & \textcolor{below60}{38.16} & \textbf{\textcolor{below60}{41.89}} &  \textcolor{above70}{75.66} & \textcolor{above70}{73.32} & \textbf{\textcolor{above70}{73.34}} \\
GPT-3.5-turbo~(\textsc{CoT})  & \textcolor{above70}{79.76} & \textcolor{above60}{65.20} & \textbf{\textcolor{above70}{70.17}} &  \textcolor{above70}{76.20} & \textcolor{above70}{76.76} & \textbf{\textcolor{above70}{75.99}}  & \textcolor{above70}{74.78} & \textcolor{below60}{37.01} & \textbf{\textcolor{below60}{41.78}} &  \textcolor{above70}{75.08} & \textcolor{above70}{73.40} & \textbf{\textcolor{above70}{73.51}} \\
GPT-4o  & \textcolor{above80}{80.51} & \textcolor{above70}{71.42} & \textbf{\textcolor{above70}{74.15}} &  \textcolor{above70}{77.51} & \textcolor{above70}{77.33} & \textbf{\textcolor{above70}{77.18}}  & \textcolor{above70}{76.02} & \textcolor{below60}{43.13} & \textbf{\textcolor{below60}{46.77}} &  \textcolor{above70}{78.81} & \textcolor{above70}{76.85} & \textbf{\textcolor{above70}{77.00}} \\
GPT-4o~(\textsc{CoT})  & \underline{\textcolor{above80}{87.02}} & \underline{\textcolor{above70}{76.47}} & \textbf{\underline{\textcolor{above70}{79.89}}} &  \textcolor{above70}{79.44} & \textcolor{above70}{79.45} & \textbf{\textcolor{above70}{79.33}}  & \textcolor{above80}{81.21} & {\textcolor{below60}{49.30}} & \textbf{{\textcolor{below60}{53.28}}} &  \textcolor{above70}{79.38} & \textcolor{above70}{77.72} & \textbf{\textcolor{above70}{77.65}} \\
QwQ-32B & \textcolor{above70}{76.96} & \textcolor{above60}{65.79} & \textcolor{above60}{\textbf{69.42}} & \textcolor{above70}{78.64} & \textcolor{above70}{78.83} & \textcolor{above70}{\textbf{78.02}} & \textcolor{above80}{86.42} & \textcolor{below60}{50.02} & \textcolor{below60}{\textbf{56.30}} & \textcolor{above80}{80.67} & \textcolor{above80}{82.82} & \textcolor{above80}{\textbf{82.15}} \\
DeepSeek-R1  & \textcolor{above80}{84.67} & \textcolor{above70}{74.82} & \textcolor{above70}{\textbf{78.25}} & \textcolor{above80}{80.04} & \textcolor{above70}{79.42} & \textcolor{above70}{\textbf{79.89}} & \textcolor{above80}{80.93} & \textcolor{below60}{47.77} & \textcolor{below60}{\textbf{52.15}} & \textcolor{above80}{80.27} & \textcolor{above80}{80.20} & \textcolor{above70}{\textbf{79.23}} \\
Grok-3-mini & \textcolor{above80}{82.41} & \textcolor{above60}{68.94} & \textcolor{above70}{\textbf{73.32}} & \textcolor{above70}{79.11} & \textcolor{above80}{80.38} & \textcolor{above80}{\textbf{80.34}} & \textcolor{above80}{87.66} & \underline{\textcolor{below60}{50.06}} & \textcolor{below60}{\textbf{56.03}} & \textcolor{above80}{81.18} & \underline{\textcolor{above80}{83.90}} & \underline{\textcolor{above80}{\textbf{82.68}}} \\
O4-mini & \textcolor{above80}{84.03} & \textcolor{above70}{71.06} & \textcolor{above70}{\textbf{74.82}} & \textcolor{above70}{79.28} & \textcolor{above80}{80.52} & \textcolor{above80}{\textbf{80.19}} & \underline{\textcolor{above80}{89.89}} & \textcolor{below60}{49.91} & \underline{\textcolor{below60}{\textbf{56.46}}} & \underline{\textcolor{above80}{81.51}} & \textcolor{above80}{83.33} & \textcolor{above80}{\textbf{82.32}} \\
\bottomrule
    \end{tabular}
    }
    \vspace{-7pt}
    \caption{Performance of different models on two text-to-KG datasets, with the highest scores in each metric \underline{underlined}.} 
    \vspace{-10pt}
    \label{tab:text-to-kg}
\end{table*}

\section{Result Analysis} \label{sec:app_result_analysis}

\subsection{Text-to-Table Generation}
As shown in Table~\ref{tab:text-to-table}, on the InstructIE dataset, the thinking models demonstrate strong performance, with Grok-3-mini achieving the highest precision of 81.82 and R1 achieving the highest recall of 78.43; however, CoT does not produce consistent improvements across all models. Regarding coherence, while the thinking models maintain excellent performance, GPT-4o achieves the highest score. Among SFT models, the R1-distilled Llama-8B exhibits the best overall performance, although it still falls short of the thinking models' capabilities. On the Struc-Bench dataset, most LLMs show strong baseline performance since the tasks primarily involve information replication rather than complex reasoning. In terms of specific metrics, Mistral-7B (SFT) achieves the highest precision score, while Llama-3.1-405B achieves the highest recall score. For coherence evaluation, GPT-4o exhibits superior capabilities. Notably, the Mistral-7B (SFT) model produces great performance gains, with improvements of 56\% in faithfulness and 34\% in coherence metrics compared to zero-shot approaches, empirically demonstrating that SFT can significantly enhance model performance on information-replication text-to-table tasks. On the Livesum dataset, both precision and recall scores are generally low due to the complex information integration and reasoning operations required by the task. O4-mini achieves exceptional performance with a faithfulness score of 83.99. Regarding coherence, all models attain relatively high scores due to the dataset's standardized table format, with Llama-3.3-70B demonstrating the best performance. The results indicate that thinking models maintain a competitive advantage on this dataset, while among SFT models, Llama-3.1-8B achieves comparable performance to larger LLMs.

\subsection{Text-to-Graph Generation}

As shown in Table~\ref{tab:text-to-kg}, our experimental results demonstrate that GPT-4o (CoT) achieves the highest faithfulness scores on the DART dataset, while Mistral-7B (SFT) shows superior coherence performance that significantly outperforms all other LLMs. The application of CoT prompting generally improves precision and recall metrics by 6-10\% across most LLMs, though it does not yield significant improvement in coherence metrics. On the InstructIE dataset, inherent dataset characteristics lead to lower recall scores overall, but QwQ-32B maintains relatively better performance in this metric, with O4-mini achieving the highest precision score of 89.45. The coherence metrics show remarkable consistency with the patterns observed on the DART dataset, suggesting stable model behavior across different evaluation benchmarks.

\begin{table}[!h]
    \centering
    \renewcommand\arraystretch{1.0}
    \scalebox{0.66}{
    \begin{tabular}{p{3.6cm}P{0.8cm}P{0.7cm}P{0.7cm}P{0.7cm}P{0.7cm}P{0.7cm}P{0.7cm}}
        \toprule
        \multirow{3}{*}{\textbf{Model}}  
        & \multicolumn{7}{c}{\textbf{Text2Chart31}} \\
        & \multicolumn{7}{c}{{\footnotesize \cite{DBLP:conf/emnlp/ZadehKKK24}}} \\
         \cmidrule(r){2-8}
         &\footnotesize \textsc{\textbf{Succ}.}& \footnotesize Prec.  & \footnotesize Rec. & \footnotesize \textsc{\textbf{Faith}}. & \footnotesize Align.  & \footnotesize Clr. &\footnotesize \textsc{\textbf{Coh}.}  \\
        \midrule
        \multicolumn{8}{l}{$\vardiamondsuit$ \emph{\textbf{SFT}}} \\
Mistral-7B & \textcolor{above70}{72.8\%} & \textcolor{above60}{64.04} & \textcolor{above60}{65.66} & \textbf{\textcolor{above60}{64.54}} &  \textcolor{above60}{63.38} & \textcolor{above60}{63.70} & \textbf{\textcolor{above60}{63.41}} \\
Llama-3.1-8B & \textcolor{above80}{87.2\%} & \textcolor{above70}{76.25} & \textcolor{above70}{79.16} & \textbf{\textcolor{above70}{77.06}} &  \textcolor{above70}{77.07} & \textcolor{above70}{77.70} & \textbf{\textcolor{above70}{77.25}} \\
R1-Distill-Llama-8B & \textcolor{below60}{47.6\%} & \textcolor{below60}{41.36} & \textcolor{below60}{42.46} & \textbf{\textcolor{below60}{41.66}} &  \textcolor{below60}{39.97} & \textcolor{below60}{41.11} & \textbf{\textcolor{below60}{40.45}} \\
\midrule
        \multicolumn{8}{l}{$\vardiamondsuit$ \emph{\textbf{Zero-Shot}}} \\
Phi-3-medium & \textcolor{above60}{68.8\%} & \textcolor{below60}{59.62} & \textcolor{above60}{62.20} & \textbf{\textcolor{above60}{60.24}} &  \textcolor{above60}{60.26} & \textcolor{above60}{60.89} & \textbf{\textcolor{above60}{60.45}} \\
Phi-3-medium~(\textsc{CoT}) & \textcolor{above60}{67.6\%} & \textcolor{below60}{56.93} & \textcolor{below60}{59.05} & \textbf{\textcolor{below60}{57.75}} &  \textcolor{below60}{58.17} & \textcolor{below60}{58.84} & \textbf{\textcolor{below60}{58.27}} \\
Mistral-7B & \textcolor{above60}{64.0\%} & \textcolor{below60}{51.72} & \textcolor{below60}{53.84} & \textbf{\textcolor{below60}{52.25}} &  \textcolor{below60}{52.03} & \textcolor{below60}{52.70} & \textbf{\textcolor{below60}{52.03}} \\
Mistral-7B~(\textsc{CoT}) & \textcolor{below60}{59.2\%} & \textcolor{below60}{44.78} & \textcolor{below60}{46.38} & \textbf{\textcolor{below60}{45.13}} &  \textcolor{below60}{46.90} & \textcolor{below60}{47.00} & \textbf{\textcolor{below60}{46.51}} \\
Mixtral-8x22B & \textcolor{above70}{72.4\%} & \textcolor{above60}{60.89} & \textcolor{above60}{63.05} & \textbf{\textcolor{above60}{61.62}} &  \textcolor{above60}{62.09} & \textcolor{above60}{62.51} & \textbf{\textcolor{above60}{62.15}} \\
Mixtral-8x22B~(\textsc{CoT}) & \textcolor{above60}{67.6\%} & \textcolor{below60}{58.23} & \textcolor{above60}{60.49} & \textbf{\textcolor{below60}{58.92}} &  \textcolor{below60}{58.61} & \textcolor{below60}{59.37} & \textbf{\textcolor{below60}{58.89}} \\
Llama-3.1-405B & \textcolor{above90}{98.0\%} & \textcolor{above80}{85.64} & \textcolor{above80}{88.00} & \textbf{\textcolor{above80}{86.49}} &  \textcolor{above80}{86.47} & \textcolor{above80}{87.26} & \textbf{\textcolor{above80}{86.62}} \\
Llama-3.1-405B~(\textsc{CoT}) & \textcolor{above90}{98.4\%} & \textcolor{above80}{86.39} & \textcolor{above80}{88.71} & \textbf{\textcolor{above80}{87.23}} &  \textcolor{above80}{86.40} & \textcolor{above80}{87.15} & \textbf{\textcolor{above80}{86.40}} \\
Llama-3.3-70B & \textcolor{above90}{98.4\%} & \textcolor{above80}{86.77} & \textcolor{above80}{89.37} & \textbf{\textcolor{above80}{87.44}} &  \textcolor{above80}{86.54} & \textcolor{above80}{88.00} & \textbf{\textcolor{above80}{87.12}} \\
Llama-3.3-70B~(\textsc{CoT}) & {\textcolor{above90}{98.8\%}} & \underline{\textcolor{above80}{88.18}} & \underline{\textcolor{above90}{90.16}} & \textbf{\underline{\textcolor{above80}{88.78}}} &  {\textcolor{above80}{87.19}} & {\textcolor{above80}{88.80}} & \textbf{{\textcolor{above80}{87.83}}} \\
GPT-3.5-turbo & \textcolor{above90}{94.4\%} & \textcolor{above80}{84.53} & \textcolor{above80}{86.23} & \textbf{\textcolor{above80}{85.17}} &  \textcolor{above80}{84.20} & \textcolor{above80}{85.63} & \textbf{\textcolor{above80}{84.79}} \\
GPT-3.5-turbo~(\textsc{CoT}) & \textcolor{above90}{93.2\%} & \textcolor{above80}{82.48} & \textcolor{above80}{84.27} & \textbf{\textcolor{above80}{83.06}} &  \textcolor{above80}{83.63} & \textcolor{above80}{84.20} & \textbf{\textcolor{above80}{83.62}} \\
GPT-4o & \textcolor{above90}{95.6\%} & \textcolor{above80}{85.16} & \textcolor{above80}{86.35} & \textbf{\textcolor{above80}{85.48}} &  \textcolor{above80}{85.91} & \textcolor{above80}{86.83} & \textbf{\textcolor{above80}{86.01}} \\
GPT-4o~(\textsc{CoT}) & \textcolor{above90}{96.0\%} & \textcolor{above80}{85.48} & \textcolor{above80}{86.82} & \textbf{\textcolor{above80}{85.99}} &  \textcolor{above80}{86.24} & \textcolor{above80}{87.26} & \textbf{\textcolor{above80}{86.63}} \\
QwQ-32B & \textcolor{above90}{98.4\%} & \textcolor{above80}{86.28} & \textcolor{above80}{88.46} & \textbf{\textcolor{above80}{86.71}} & \textcolor{above80}{86.61} & \textcolor{above80}{88.04} & \textbf{\textcolor{above80}{86.94}} \\
DeepSeek-R1 & \textcolor{above90}{95.6\%} & \textcolor{above80}{85.36} & \textcolor{above80}{87.16} & \textbf{\textcolor{above80}{85.79}} &  \textcolor{above80}{85.71} & \textcolor{above80}{87.64} & \textbf{\textcolor{above80}{86.56}} \\
Grok-3-mini & \underline{\textcolor{above90}{99.2\%}} & \textcolor{above80}{87.44} & \textcolor{above80}{89.19} & \textbf{\textcolor{above80}{88.05}} & \underline{\textcolor{above80}{88.34}} & \underline{\textcolor{above80}{89.65}} & \underline{\textbf{\textcolor{above80}{88.74}}} \\
O4-mini & \textcolor{above90}{95.6\%} & \textcolor{above80}{84.92} & \textcolor{above80}{87.30} & \textbf{\textcolor{above80}{85.56}} & \textcolor{above80}{85.84} & \textcolor{above80}{87.16} & \textbf{\textcolor{above80}{86.17}} \\
\bottomrule
    \end{tabular}}
    \vspace{-7pt}
    \caption{Performance of different models on text-to-chart dataset, with the highest scores in each metric \underline{underlined}.} 
    \vspace{-10pt}
    \label{tab:text-to-chart}
\end{table}

\subsection{Text-to-Chart Generation}

As shown in Table~\ref{tab:text-to-chart}, in addition to the aforementioned metrics, we also evaluate the success rate of the generated Python code executing correctly to produce the desired images, denoted as \textsc{Succ}. It is observed that the majority of LLMs achieve a success rate of over 95\%. Among them, Grok-3-mini demonstrates the highest success rate, reaching 99.2\%. In terms of precision an recall, most LLMs achieve scores above 80, with Llama-3.3-70B (CoT) performing the highest, scoring 88.18 and 90.16, respectively. For alignment and clarity, Grok-3-mini exhibits the best performance, achieving scores of 88.34 and 89.65, respectively. Among the SFT models, Llama-3.1-8B demonstrated the highest overall performance. Mistral-7B achieves significant improvements compared to its zero-shot counterpart, with a 13.8\% increase in success rate, 23.5\% and 21.9\% enhancements in faithfulness and coherence scores, respectively, showing the effectiveness of SFT in model optimization.

\section{Case Study} \label{app:case_study}

This section illustrates five representative failure cases of traditional evaluation metrics through systematic analysis. We empirically demonstrate scenarios where traditional metrics (BERTScore, ROUGE, etc.) contradict human judgment while our framework correctly identifies superior outputs. Each case dissects specific error types and metric failures, providing granular evidence for our approach's validity across diverse tasks. To reflect the scoring process of \textit{Content Faithfulness} and \textit{Structure Coherence}, we append the intermediate reasoning steps from the LLM as a brief justification following the output for each case.

\paragraph{Case 1 (Figure~\ref{fig:case1})} \label{p:case1}
It is notable that the table generated by GPT-3.5 lacks two critical rows of information, and its single extracted row contains multiple errors compared to the ground truth. Conversely, the table produced by R1 contains comprehensive information, fully capturing all details present in the ground truth while additionally extracting more complete information. Despite R1's table being demonstrably superior in this context, both BERTScore and ROUGE metrics indicate GPT-3.5's output is preferable.

\paragraph{Case 2 (Figure~\ref{fig:case2})} \label{p:case2}
The table generated by GPT-3.5 introduced erroneous information in several cells (e.g., misrepresenting reactants and products). In contrast, R1's extracted table maintained complete accuracy in all key information. Paradoxically, BERTScore and ROUGE metrics significantly favor the former.

\paragraph{Case 3 (Figure~\ref{fig:case3})} \label{p:case3}
GPT-3.5's table exhibits numerical inaccuracies across multiple cells, with an error rate 60\% higher than R1's table. Nevertheless, BERTScore metrics consistently rate GPT-3.5's output as superior.

\paragraph{Case 4 (Figure~\ref{fig:case4})} \label{p:case4}
GPT-3.5's edge information contains instances where predicates and objects are entirely confused, accompanied by structural disarray. Conversely, R1 extracts a structurally correct graph. Contradictorily, both BERTScore and ROUGE metrics indicate GPT-3.5's graph extraction performed better.

\begin{figure*}
    \centering
\begin{tcolorbox}[title={Case 5: Instance \#0208113110 from Text2Chart31}, colback = cBlue_2!10, colframe = cBlue_5!80,  coltitle=white,fonttitle=\bfseries\small,fontupper=\scriptsize,fontlower=\scriptsize]

\begin{center}

{\small\textbf{Comparison of Evaluation Metrics}}

\begin{tabular}{L{2cm}P{1cm}P{1cm}P{2.5cm}P{2.5cm}}
\toprule
    \textbf{Output} & \textbf{METEOR} & \textbf{CodeBLEU} & \textbf{Content Faithfulness} & \textbf{Structure Coherence}  \\
    \midrule
    GPT-3.5-turbo  & \textbf{0.4480} &  \textbf{0.3701} &  74.67 & 69.64 \\  
    DeepSeek-R1  & 0.3148 &  0.2263 &  \textbf{100.00} & \textbf{72.41} \\  
    \bottomrule
\end{tabular}

\end{center}
\tcblower
\begin{center}

{\small\textbf{Ground Truth}}

\includegraphics[width=3cm]{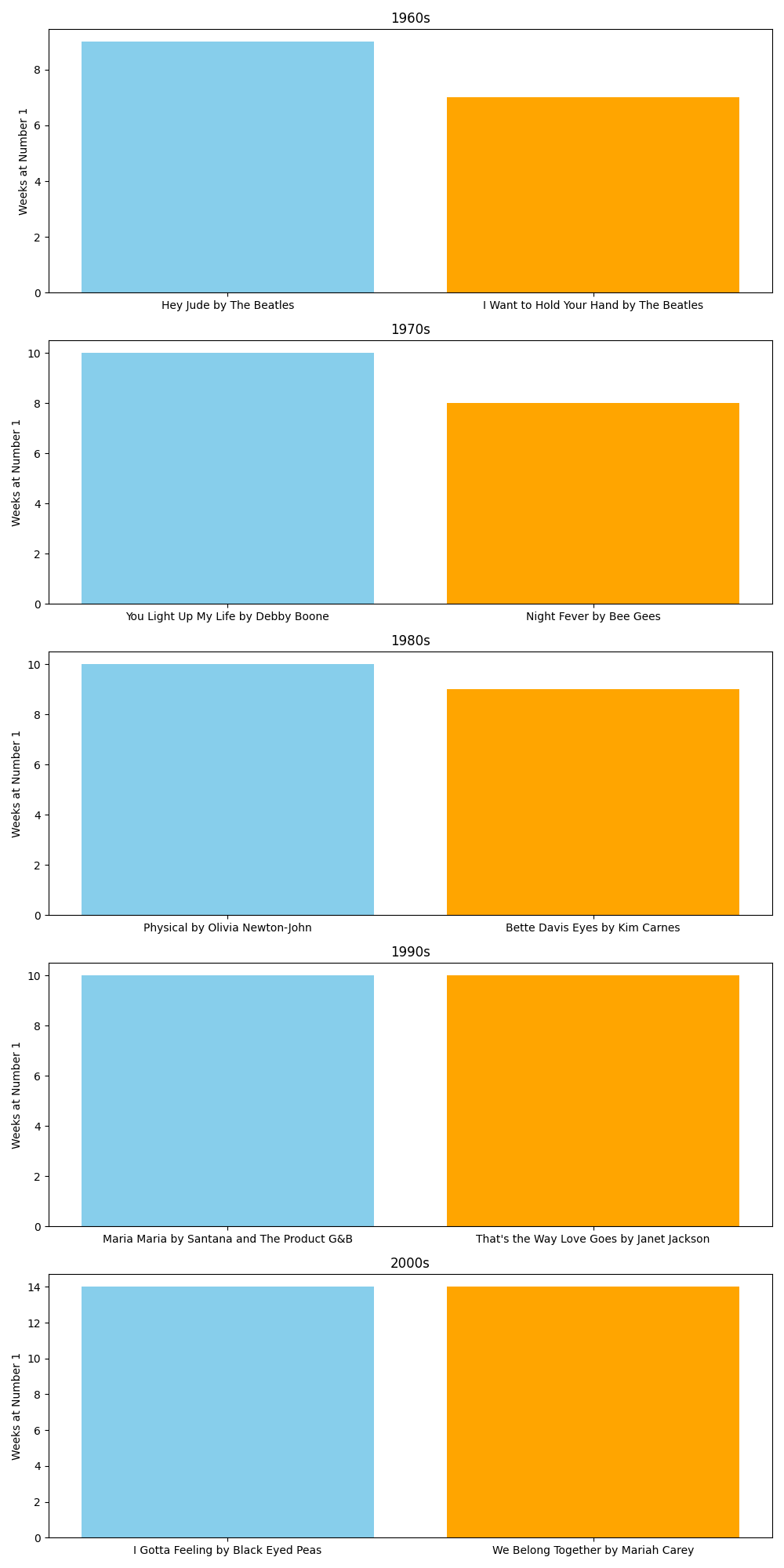}

\vspace{0.1in}

\rule{\linewidth}{0.5pt} 
\vspace{0.01in}

{\small\textbf{GPT-3.5-turbo Output}}

\includegraphics[width=5cm]{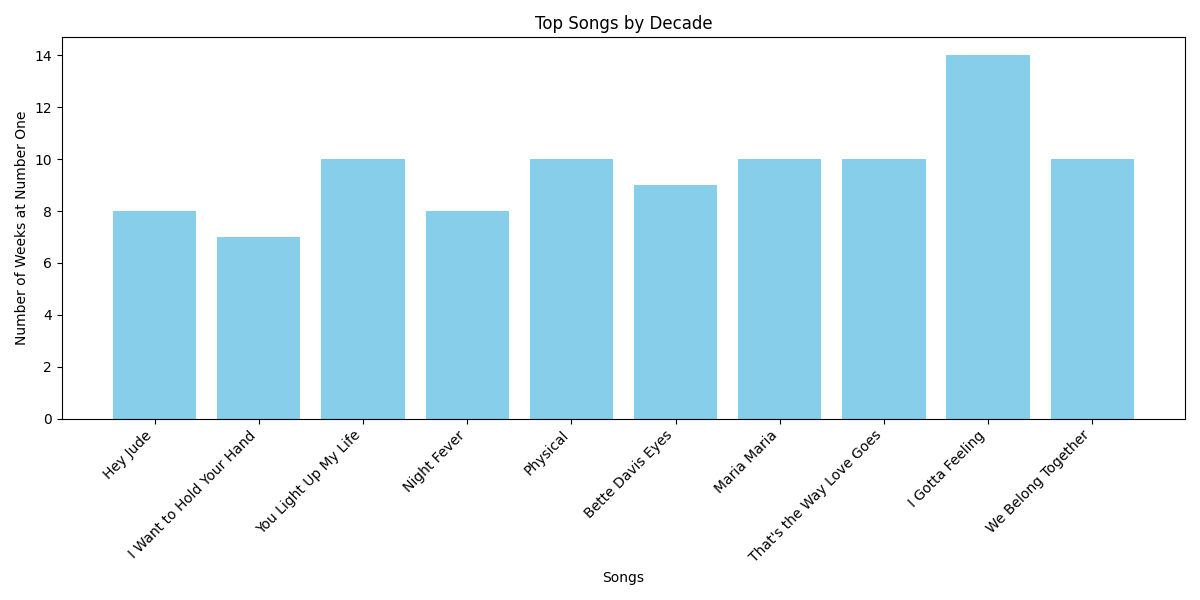}

\begin{verbatim}
**Brief Justification**
   - **Precision**: The LLM-generated chart has discrepancies in the number of weeks for some songs compared to
      the ground truth. For example, "Hey Jude" is shown with 8 weeks instead of 9, and "I Gotta Feeling" with
      14 weeks instead of 10.
   - **Recall**: The generated chart includes all songs and their respective weeks, but the values are not always
      accurate, affecting recall.
   - **Alignment**: The chart type is consistent, but the LLM-generated chart lacks the decade grouping present
      in the ground truth, which is a significant structural difference.
   - **Clarity**: The generated chart is clear but lacks the decade separation, which could lead to confusion about
      the time periods of the songs.
\end{verbatim} 
\rule{\linewidth}{0.5pt} %
\vspace{0.01in}

{\small\textbf{DeepSeek-R1 Output}}

\includegraphics[width=5cm]{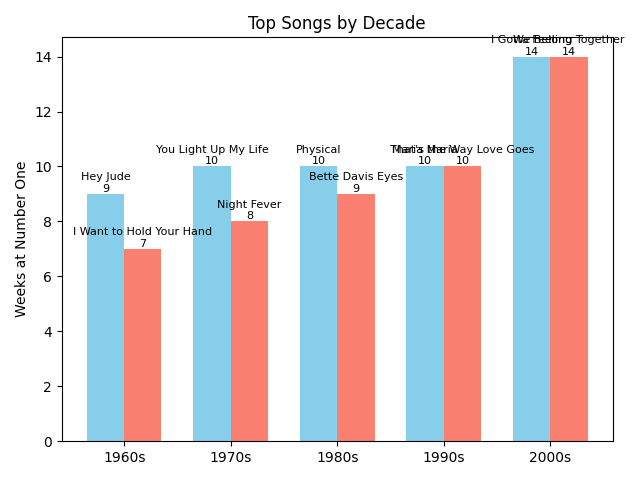}

\begin{verbatim}
**Brief Justification**
   - **Precision/Recall**: All correct.
   - **Alignment**: The chart type is consistent (bar chart), but the structure differs. The ground truth uses sepa-
      rate subplots for each decade, while the generated chart combines them into one. This structural difference impacts
      the score.
   - **Clarity**: The generated chart is clear and logical in its presentation, but the overlapping labels and combined
      decades create some confusion, affecting the semantic score.
\end{verbatim} 

\end{center}
\end{tcolorbox}

    \caption{Misalignment of Traditional Metrics and Our Framework: Case 5}
    \label{fig:case5}
\end{figure*}

\paragraph{Case 5 (Figure~\ref{fig:case5})} \label{p:case5}
GPT-3.5's generated chart contains numerical errors and fails to reflect logical pairing in its layout. R1's chart, however, maintains numerical accuracy and replicates the ground truth's logical structure. Despite this qualitative advantage, METEOR and CodeBLEU metrics favor GPT-3.5's output.

\begin{figure*}[b]
    \centering
\begin{tcolorbox}[title={Case 1: Instance \#068 from InstructIE}, colback = cBlue_2!10, colframe = cBlue_5!80,  coltitle=white,fonttitle=\bfseries\small,fontupper=\scriptsize,fontlower=\scriptsize]

\begin{center}
{\small\textbf{Comparison of Evaluation Metrics}}

\begin{tabular}{L{2cm}P{1cm}P{1cm}P{2.5cm}P{2.5cm}}
\toprule
    \textbf{Output} & \textbf{BERTScore} & \textbf{ROUGE} & \textbf{Content Faithfulness} & \textbf{Structure Coherence}  \\
    \midrule
    GPT-3.5-turbo  & \textbf{0.8602} &  \textbf{0.7052} &  40.00 & 64.62 \\  
    DeepSeek-R1  & 0.8364 &  0.5974 &  \textbf{75.00} & \textbf{80.00} \\  
    \bottomrule
\end{tabular}

\end{center}
\tcblower
\begin{center}
{\small\textbf{Ground Truth}}

\begin{tabular}{|m{2.5cm}|m{2.5cm}|m{3cm}|m{2cm}|m{2.5cm}|}
\hline
\textbf{Number of Bedrooms} & \textbf{Number of Bathrooms} & \textbf{Special Rooms/Spaces} & \textbf{Floor Material} & \textbf{Outdoor Spaces} \\
\hline
3 & 2.5 & Breakfast nook, playroom & Hardwood & Fenced backyard, deck, patio \\
\hline
3 & N/A & Den, indoor rock garden & New flooring & Rear balconies \\
\hline
5 & 3.5 & Family room, playroom, laundry & Wood & Large back patio, storage shed, fenced yard \\
\hline
\end{tabular}

\vspace{0.1in}

\rule{\linewidth}{0.5pt} %
\vspace{0.01in}

{\small\textbf{GPT-3.5-turbo Output}}

\begin{tabular}{|m{2.5cm}|m{2.5cm}|m{3cm}|m{2cm}|m{2.5cm}|}
\hline
\textbf{Number of Bedrooms} & \textbf{Number of Bathrooms} & \textbf{Special Rooms/Spaces} & \textbf{Floor Material} & \textbf{Outdoor Spaces} \\
\hline
3 & 2.5 & Playroom, Master Suite, Den & Hardwood floors, New flooring & Deck, Patio, Fenced backyard, Front patio area \\
\hline
\end{tabular}

\begin{verbatim}
**Brief Justification**:
   - **Precision**: 3/5 cells conflicting (Special Rooms, Floor Material, Outdoor Spaces) → 40.  
   - **Recall**: 3/5 ground truth cells missing/incorrect (Breakfast nook, Hardwood, exact Outdoor Spaces) → 40.  
   - **Alignment**: Missing two rows (ground truth has 3 rows, LLM has 1) → Noticeable Differences (70).  
   - **Clarity**: Merged data from multiple sources into one row creates ambiguity → Significant Issues (60). 
\end{verbatim} 
\rule{\linewidth}{0.5pt} %
\vspace{0.01in}

{\small\textbf{DeepSeek-R1 Output}}

\begin{tabular}{|m{0.2cm}|m{1.1cm}|m{1.1cm}|m{5.5cm}|m{2cm}|m{2cm}|}
\hline
\textbf{\#} & \textbf{Bedrooms} & \textbf{Bathrooms} & \textbf{Special Rooms/Spaces} & \textbf{Floor Material} & \textbf{Outdoor Spaces} \\
\hline
1 & 3 & 2.5 & Pantry, kitchen island, breakfast nook, master suite (walk-in closets), playroom, laundry area, garage/workshop & Hardwood & Fenced backyard, deck, patio, front patio \\
\hline
2 & 3 & N/A & Two-story living room, conversation pit, indoor rock garden, den, updated kitchen & New flooring & Rear balconies, golf course views, lake views \\
\hline
3 & 5 & 3.5 & Kitchen island, pantry, laundry room, primary suite (walk-in closet), finished basement (family room, playroom) & Wood floors & Back patio, storage shed with loft, fenced yard \\
\hline
\end{tabular}

\begin{verbatim}
**Brief Justification**  
   - **Precision**: LLM-added details (e.g., extra rooms/outdoor features) not in GT create 6/15 conflicts (60).  
   - **Recall**: All GT cells are captured despite extra LLM additions (100).  
   - **Alignment**: Extra row numbers and header wording deviations (80).  
   - **Clarity**: Clear headers but redundant cell details (80). 
\end{verbatim} 

\end{center}
\end{tcolorbox}

    \caption{Misalignment of Traditional Metrics and Our Framework: Case 1}
    \label{fig:case1}
\end{figure*}

\begin{figure*}
    \centering
\begin{tcolorbox}[title={Case 2: Instance \#127 from InstructIE}, colback = cBlue_2!10, colframe = cBlue_5!80,  coltitle=white,fonttitle=\bfseries\small,fontupper=\scriptsize,fontlower=\scriptsize]

\begin{center}

{\small\textbf{Comparison of Evaluation Metrics}}

\begin{tabular}{L{2cm}P{1cm}P{1cm}P{2.5cm}P{2.5cm}}
\toprule
    \textbf{Output} & \textbf{BERTScore} & \textbf{ROUGE} & \textbf{Content Faithfulness} & \textbf{Structure Coherence}  \\
    \midrule
    GPT-3.5-turbo  & \textbf{0.9229} &  \textbf{0.9012} &  61.46 & 67.41 \\  
    DeepSeek-R1  & 0.7231 &  0.4086 &  \textbf{100.00} & \textbf{76.77} \\  
    \bottomrule
\end{tabular}
\end{center}
\tcblower
\begin{center}

{\small\textbf{Ground Truth}}

\begin{tabular}{|m{4cm}|m{2cm}|m{4cm}|m{2cm}|}
\hline
\textbf{Reactants} & \textbf{Solvent} & \textbf{Product} & \textbf{Yield} \\
\hline
CyPBn-Cy, NiCl\textsubscript{2}(DME) & THF & (CyPBn-Cy)NiCl\textsubscript{2} & 68\% \\
\hline
(CyPBn-Cy)NiCl\textsubscript{2}, (o-tol)MgCl & THF & (CyPBn-Cy)Ni(o-tol)Cl & 95\% \\
\hline
\end{tabular}

\vspace{0.1in}

\rule{\linewidth}{0.5pt} %
\vspace{0.01in}

{\small\textbf{GPT-3.5-turbo Output}}

\begin{tabular}{|m{4cm}|m{4cm}|m{1cm}|m{1cm}|m{3cm}|}
\hline
\textbf{Reactants} & \textbf{Products} & \textbf{Solvent} & \textbf{Yield} & \textbf{Analysis Method} \\
\hline
CyPBn-Cy, NiCl\textsubscript{2}(DME), THF & (CyPBn-Cy)NiCl\textsubscript{2}, dichloromethane & THF & 68\% & Elemental, NMR, X-ray \\
\hline
(CyPBn-Cy)NiCl\textsubscript{2}, (o-tol)MgCl, THF & (CyPBn-Cy)Ni(o-tol)Cl & THF & 95\% & -- \\
\hline
\end{tabular}

\begin{verbatim}
**Brief Justification**  
   - **Precision**: 4/10 cells incorrect (extra "dichloromethane" in Product, Analysis Method column invalid, THF 
                    mislabeled as Reactant).  
   - **Recall**: 3/8 GT cells missing/incorrect (Reactants misrepresented in both rows, Product error in row 1).  
   - **Alignment**: Added Analysis Method column, Products vs Product header mismatch (Noticeable Differences).  
   - **Clarity**: Ambiguous Reactants (THF included redundantly), redundant Analysis Method (Significant Issues).
\end{verbatim} 
\rule{\linewidth}{0.5pt} %
\vspace{0.01in}

{\small\textbf{DeepSeek-R1 Output}}

\begin{tabular}{|m{1cm}|m{2cm}|m{3cm}|m{1cm}|m{2cm}|m{0.5cm}|m{2.5cm}|}
\hline
\textbf{Reaction Equation} & \textbf{Reactants} & \textbf{Products} & \textbf{Solvent} & \textbf{Workup/Isolation} & \textbf{Yield} & \textbf{Analytical Data} \\
\hline
eq1 & CyPBn-Cy, NiCl\textsubscript{2}(DME) & (CyPBn-Cy)NiCl\textsubscript{2} (CH\textsubscript{2}Cl\textsubscript{2} solvate) & THF & Dichloromethane solvate & 68\% & Elemental analysis, NMR, X-ray crystallography \\
\hline
eq2 & (CyPBn-Cy)NiCl\textsubscript{2}, (o-tol)MgCl & (CyPBn-Cy)Ni(o-tol)Cl & THF & Isolated as pure solid & 95\% & Elemental analysis \\
\hline
\end{tabular}

\begin{verbatim}
- **Brief Justification**  
   - **Precision**: All corresponding cells (Reactants, Solvent, Product, Yield) match GT and source text. 
                    Extra columns (Workup/Isolation, Analytical Data) align with source and do not conflict.  
   - **Recall**: All GT cells are captured in the generated table.  
   - **Alignment**: Added columns (Reaction Equation, Workup/Isolation, Analytical Data) deviate from GT schema.  
   - **Clarity**: Clear headers and data, but redundant columns introduce minor deviations from GT focus.
\end{verbatim} 

\end{center}
\end{tcolorbox}

    \caption{Misalignment of Traditional Metrics and Our Framework: Case 2}
    \label{fig:case2}
\end{figure*}

\begin{figure*}
    \centering
\begin{tcolorbox}[title={Case 3: Instance \#42842344 from LiveSum}, colback = cBlue_2!10, colframe = cBlue_5!80,  coltitle=white,fonttitle=\bfseries\small,fontupper=\scriptsize,fontlower=\scriptsize]

\begin{center}
{\small\textbf{Comparison of Evaluation Metrics}}

\begin{tabular}{L{2cm}P{1cm}P{1cm}P{2.5cm}P{2.5cm}}
\toprule
    \textbf{Output} & \textbf{BERTScore} & \textbf{ROUGE} & \textbf{Content Faithfulness} & \textbf{Structure Coherence}  \\
    \midrule
    GPT-3.5-turbo  & \textbf{0.9303} &  {0.7750} &  50.00 & 100.00 \\  
    DeepSeek-R1  & 0.9294 &  0.7750 &  \textbf{69.00} & 100.00 \\  
    \bottomrule
\end{tabular}
\end{center}
\tcblower
\begin{center}
{\small\textbf{Ground Truth}}

\begin{tabular}{|m{1.5cm}|m{0.8cm}|m{0.8cm}|m{0.8cm}|m{1.5cm}|m{1.5cm}|m{1.5cm}|m{1.5cm}|m{1cm}|}
\hline
\textbf{Team} & \textbf{Goals} & \textbf{Shots} & \textbf{Fouls} & \textbf{Yellow Cards} & \textbf{Red Cards} & \textbf{Corner Kicks} & \textbf{Free Kicks} & \textbf{Offsides} \\
\hline
Home Team & 5 & 15 & 8 & 2 & 0 & 4 & 8 & 1 \\
\hline
Away Team & 1 & 10 & 8 & 0 & 0 & 6 & 8 & 3 \\
\hline
\end{tabular}

\vspace{0.1in}

\rule{\linewidth}{0.5pt} %
\vspace{0.01in}

{\small\textbf{GPT-3.5-turbo Output}}

\begin{tabular}{|m{1.5cm}|m{0.8cm}|m{0.8cm}|m{0.8cm}|m{1.5cm}|m{1.5cm}|m{1.5cm}|m{1.5cm}|m{1cm}|}
\hline
\textbf{} & \textbf{Goals} & \textbf{Shots} & \textbf{Fouls} & \textbf{Yellow Cards} & \textbf{Red Cards} & \textbf{Corner Kicks} & \textbf{Free Kicks} & \textbf{Offsides} \\
\hline
Home Team & 5 & 12 & 6 & 1 & 0 & 5 & 4 & 1 \\
\hline
Away Team & 1 & 8 & 7 & 0 & 0 & 4 & 6 & 3 \\
\hline
\end{tabular}

\begin{verbatim}
**Brief Justification**  
   - **Precision**: 8/16 cells have errors (numerical mismatches), leading to 50% error rate.  
   - **Recall**: Same 8 missing/correct GT cells, 50% recall.  
   - **Alignment**: Columns/headers match; row order irrelevant per criteria (no penalty).  
   - **Clarity**: Headers clear, data unambiguous despite errors. 
\end{verbatim} 
\rule{\linewidth}{0.5pt} %
\vspace{0.01in}

{\small\textbf{DeepSeek-R1 Output}}

\begin{tabular}{|m{2cm}|m{2cm}|m{2cm}|}
\hline
\textbf{Stat} & \textbf{Home Team} & \textbf{Away Team} \\
\hline
Goals & 5 & 1 \\
\hline
Shots & 12 & 8 \\
\hline
Fouls & 8 & 12 \\
\hline
Yellow Cards & 2 & 0 \\
\hline
Red Cards & 0 & 0 \\
\hline
Corner Kicks & 5 & 5 \\
\hline
Free Kicks & 12 & 8 \\
\hline
Offsides & 1 & 3 \\
\hline
\end{tabular}

\begin{verbatim}
**Brief Justification**
   - **Precision/Recall**: 5 cell errors (Shots, Fouls, Corner Kicks, Free Kicks).
   - **Alignment**: Transposed correctly without penalties.
   - **Clarity**: Clear headers and layout despite data inaccuracies.
\end{verbatim} 

\end{center}
\end{tcolorbox}

    \caption{Misalignment of Traditional Metrics and Our Framework: Case 3}
    \label{fig:case3}
\end{figure*}

\begin{figure*}
    \centering
\begin{tcolorbox}[title={Case 4: Instance \#048 from DART}, colback = cBlue_2!10, colframe = cBlue_5!80,  coltitle=white,fonttitle=\bfseries\small,fontupper=\scriptsize,fontlower=\scriptsize]

\begin{center}

{\small\textbf{Comparison of Evaluation Metrics}}

\begin{tabular}{L{2cm}P{1cm}P{1cm}P{2.5cm}P{2.5cm}}
\toprule
    \textbf{Output} & \textbf{BERTScore} & \textbf{ROUGE} & \textbf{Content Faithfulness} & \textbf{Structure Coherence}  \\
    \midrule
    GPT-3.5-turbo  & \textbf{0.7165} &  \textbf{0.6465} &  40.48 & 74.67 \\  
    DeepSeek-R1  & 0.7052 &  0.6166 &  \textbf{100.00} & \textbf{80.00} \\  
    \bottomrule
\end{tabular}

\end{center}
\tcblower
\begin{center}

{\small\textbf{Ground Truth}}

\begin{tabular}{|m{4cm}|m{4cm}|m{5cm}|}
\hline
\textbf{Subject} & \textbf{Predicate} & \textbf{Object} \\
\hline
The Vaults & eatType & pub \\
\hline
The Vaults & food & Japanese \\
\hline
The Vaults & priceRange & £20--25 \\
\hline
The Vaults & customer rating & high \\
\hline
The Vaults & area & city centre \\
\hline
The Vaults & familyFriendly & yes \\
\hline
The Vaults & near & Rainbow Vegetarian Café \\
\hline
\end{tabular}

\vspace{0.1in}

\rule{\linewidth}{0.5pt} %
\vspace{0.01in}

{\small\textbf{GPT-3.5-turbo Output}}

\begin{tabular}{|m{4cm}|m{4cm}|m{5cm}|}
\hline
\textbf{Subject} & \textbf{Predicate} & \textbf{Object} \\
\hline
The Vaults & has price range & £20--25 \\
\hline
The Vaults & is child friendly & \\
\hline
The Vaults & is a Japanese pub & \\
\hline
The Vaults & is near & Rainbow Vegetarian Café \\
\hline
The Vaults & is located in & city centre \\
\hline
The Vaults & is highly rated & \\
\hline
\end{tabular}

\begin{verbatim}
**Brief Justification**  
   - **Precision**: 2/6 triples conflict (missing "yes" and "high" values).  
   - **Recall**: 5/7 ground truth triples missing/incorrect (combined or incomplete).  
   - **Structural**: Missing objects in 2 triples and merged triples, parsable but flawed.  
   - **Semantic**: Non-standard predicates (e.g., "is child friendly") but core meaning preserved.   
\end{verbatim} 
\rule{\linewidth}{0.5pt} %
\vspace{0.01in}

{\small\textbf{DeepSeek-R1 Output}}

\begin{tabular}{|m{4cm}|m{4cm}|m{5cm}|}
\hline
\textbf{Subject} & \textbf{Relation} & \textbf{Object} \\
\hline
The Vaults & is & child friendly \\
\hline
The Vaults & is a & highly rated Japanese pub \\
\hline
The Vaults & has price range & £20--25 \\
\hline
The Vaults & is located in & city centre \\
\hline
The Vaults & is near & Rainbow Vegetarian Café \\
\hline
\end{tabular}

\begin{verbatim}
**Brief Justification**
   All ground truth information is captured without conflicts, but relations use non-standard phrasing (e.g., "is a" 
   vs "eatType"). Structural/semantic scores reflect schema deviations and non-standard predicates.
\end{verbatim} 

\end{center}
\end{tcolorbox}

    \caption{Misalignment of Traditional Metrics and Our Framework: Case 4}
    \label{fig:case4}
\end{figure*}

\end{document}